\documentclass[final,5p,times,twocolumn]{elsarticle}

\usepackage{booktabs}
\usepackage{adjustbox}
\usepackage{multirow}
\usepackage{xcolor}
\usepackage{colortbl}
\usepackage{amsmath,amssymb}
\usepackage{enumitem}
\usepackage{array}
\usepackage{microtype}

\usepackage{hyperref}
\newcommand{\codeurl}{\url{https://github.com/medalimoussa/FLEX-LBP}}
\hypersetup{colorlinks=true,linkcolor=blue!45!black,citecolor=blue!45!black,urlcolor=blue!45!black,
  pdftitle={FLEX-LBP: Learning Local Binary Patterns with Exact Binary Inference for Texture Classification},
  pdfauthor={Mohamed Alimoussa, Amal El Kaid}}

\graphicspath{{figures/}}


\setlist{nosep,leftmargin=1.5em}

\newcommand{\codefigwidth}{\linewidth} %

\journal{Expert Systems with Applications}
\biboptions{authoryear}

\begin{document}

\begin{frontmatter}

\title{FLEX-LBP: Learning Local Binary Patterns with Exact Binary Inference for Texture Classification}

\author[a]{Mohamed Alimoussa\corref{cor1}}
\ead{malimoussa@insa-toulouse.fr}
\author[b]{Amal El Kaid}
\cortext[cor1]{Corresponding author.}
\address[a]{Institut Cl\'ement Ader (ICA), Universit\'e de Toulouse, Toulouse, 31000, France}
\address[b]{LIMOS, Universit\'e Clermont Auvergne, Clermont-Ferrand, France}

\begin{abstract}
Local Binary Patterns (LBP) remain among the most widely used texture descriptors, from industrial inspection to medical imaging.
Yet their radius, sampling angles, color transform and thresholds are fixed by hand for each dataset.
We present FLEX-LBP, which learns these parameters by backpropagation while every forward pass computes exact binary codes and hard histogram counts.
A straight-through estimator acts only in the backward pass, so the deployed descriptor is an interpretable hard-binary LBP with 90 learnable parameters and unchanged inference cost.
In a controlled ablation on six color texture benchmarks, the component worth learning is dataset-dependent.
One configuration, applied unchanged everywhere, learns the geometry and thresholds with the color transform frozen.
It improves the nearest-neighbor accuracy of the identical fixed descriptor by 3.55 points on average over five seeds, significantly on five of six benchmarks and never significantly worse.
With a linear discriminant readout, it gains 0.9 points; the fully learned descriptor gains 2.7 and has the highest mean.
FLEX-LBP is more accurate than five classical LBP-family descriptors, PCANet and a neural LBP on five of six benchmarks.
Its mean nearest-neighbor accuracy exceeds that of frozen ImageNet ResNet-50 features (23.5 million parameters).
On the Kather colorectal cancer histology benchmark, the fully learned descriptor reaches 93.8\%, significantly above the best classical descriptor (Completed LBP, 92.8\%).
The learned radii adapt in directions that recur across seeds.
Noise drowns single-pixel comparisons and illuminant change flips inter-channel ones; median sampling or gray-world normalization restores 96--97\%.
The code is available as a drop-in PyTorch layer.
\end{abstract}


\end{frontmatter}

\section{Introduction}
\label{sec:intro}

Local Binary Patterns (LBP)~\citep{ojala2002} are among the most widely used texture descriptors, with applications ranging from industrial surface inspection~\citep{maenpaa2003realtime} and medical image analysis~\citep{nanni2010medical} to remote sensing~\citep{li2015hyperspectral} and face recognition~\citep{ahonen2006face}; \citet{pietikainen2011book} review many more.
An LBP operator compares each pixel with neighbors on a circle around it, packs the comparison bits into an integer code and describes the image by the histogram of these codes.
LBP are cheap to compute, invariant to monotonic gray-level changes and geometrically interpretable.
These properties matter most when a task offers only a few hundred to a few thousand training images and the descriptor must be auditable.
Large pretrained networks are then not an obvious choice: they bring millions of parameters learned on other data, and their features cannot be read off.

The weakness of LBP is configuration: the classical operator fixes its radius, angular sampling directions, color-space transform, comparison threshold and intensity quantization before any data is seen.
Each new material, imaging setup or dataset therefore requires an expert to re-tune these choices by grid search, and the optimum does not transfer.
In our ablation, a change that improves one benchmark by several points degrades another by as much (Section~\ref{sec:exp:results}).
We address this tuning step and ask three questions.
(Q1)~Which components of the LBP pipeline are worth learning from data, and does the answer depend on the dataset?
(Q2)~How much of the gain over a classical LBP comes from learning, and how much from descriptor postprocessing and the readout?
(Q3)~Do the learned parameters remain interpretable?

Prior work approaches the problem from two sides.
Learned methods such as Binarized Statistical Image Features (BSIF)~\citep{kannala2012bsif} and Local Binary Convolutional Neural Networks~\citep{juefei2017lbcnn} replace the radius, angle and color parameters of LBP with generic learned filters, or embed binary comparisons in deep networks.
Hand-designed extensions such as Completed LBP (CLBP)~\citep{guo2010clbp}, Robust LBP (RLBP)~\citep{chen2013rlbp} and Median Robust Extended LBP (MRELBP)~\citep{liu2016mrelbp} show that the design space matters but do not learn it.
Neural LBP variants come closest: they learn sampling positions~\citep{jin2018lbpnet} or relaxed thresholds~\citep{peeples2024nehd} inside deep networks.
Their deployed feature, however, is no longer a histogram LBP whose radius, angles, thresholds and color basis can be read off (Section~\ref{sec:related}).

We introduce FLEX-LBP (Feature Learning with EXact Local Binary Patterns), which keeps the classical multi-scale histogram descriptor, computed by three parallel branches.
As in opponent color LBP (OCLBP)~\citep{maenpaa2004classification}, each branch compares the center pixel in one color channel with its neighbors in the same or another channel.
Following CLBP, each channel pair yields a sign and a magnitude histogram, which form the sign-plus-magnitude LBP (SM-LBP) descriptor of the branch.
FLEX-LBP learns five components by backpropagation: (i)~a per-branch $3{\times}3$ color transform; (ii)~radii bounded to scale-specific ranges; (iii)~ordered sampling angles that cover the full circle; (iv)~per-direction comparison thresholds that generalize the zero threshold of classical LBP, in the spirit of Local Ternary Patterns (LTP)~\citep{tan2010ltp}; and (v)~the number of intensity quantization levels.

Binary comparison and histogram binning are not differentiable, so we train with a straight-through estimator (STE)~\citep{bengio2013ste}.
Every forward pass, in training as at test time, computes exact binary codes and hard histogram counts, and surrogate derivatives act only in the backward pass.

We evaluate FLEX-LBP on six color texture benchmarks: NewBarkTex~\citep{porebski2014newbarktex}, KTH-TIPS2-b~\citep{caputo2005class}, Outex~TC13 and TC31~\citep{ojala2002outex,lille2022extendedoutex}, Parquet~\citep{bianconi2013parquet} and USPtex~\citep{backes2012color}.
Fixed and learned descriptors receive the same postprocessing (uniform-pattern projection and power normalization~\citep{perronnin2010fisher}) and are classified by standardized one-nearest-neighbor (1-NN) matching.
Under the same protocol, we compare FLEX-LBP with five classical and two learned LBP-family descriptors, and in accuracy and cost with frozen ImageNet ResNet-50 features~\citep{he2016resnet}.
We also analyze two Outex suites with photometric shifts (TC32, additive noise; TC34, illuminant change) on which the color descriptor fails.
Finally, we apply FLEX-LBP to the colorectal cancer histology benchmark of \citet{kather2016multi}.
To our knowledge, FLEX-LBP is the first texture descriptor that learns its sampling radii and angles, its comparison thresholds and its color transform from labeled data while the deployed operator keeps exact binary codes and histogram counts.

Our contributions are as follows:
\begin{enumerate}
  \item We propose FLEX-LBP, an LBP whose 90 stored parameters can be learned by backpropagation while every forward pass stays exactly binary. The deployed descriptor is thus a hard-binary SM-LBP with $3{,}186$ dimensions after postprocessing and the inference cost of the fixed operator (Section~\ref{sec:method}).
  \item We report a controlled component ablation with paired significance tests over training seeds. The best learned component depends on the dataset, but one configuration, Geometry, which learns the sampling geometry and the thresholds with the color transform frozen, is applied unchanged to every benchmark. Over five seeds, it improves the 1-NN accuracy of the identical fixed descriptor by $+3.55$ percentage points (pp) on average. The gain is significant on five of the six main benchmarks, and the learned descriptor is never significantly worse. It also improves the fixed descriptor on histology, a benchmark added after it was chosen (Sections~\ref{sec:exp:results}--\ref{sec:exp:significance} and~\ref{sec:exp:kather}).
  \item We separate the contributions of postprocessing, readout and learning. On the six main benchmarks, postprocessing adds $+2.02$\,pp on average to the fixed descriptor, and learning the geometry adds $+3.55$\,pp on top of it. With a supervised linear discriminant analysis (LDA) readout instead of 1-NN, the gain of the learned geometry falls to $+0.92$\,pp, whereas the fully learned descriptor keeps $+2.68$\,pp. LDA also lifts the fixed descriptor above the learned geometry read out by 1-NN on four of the six main benchmarks (Sections~\ref{sec:exp:postprocessing} and~\ref{sec:analysis:mismatch}).
  \item With the geometry learned, FLEX-LBP is more accurate than the strongest classical descriptor on five of the six main benchmarks. Averaged over seeds, it is also more accurate than the two learned ones, PCANet~\citep{chan2015pcanet} and the neural LBP (NLBP) of \citet{peeples2024nehd}, on all six. With LDA on every descriptor, FLEX-LBP is again ahead on five of the six, but MRELBP then leads on KTH-TIPS2. Fully learned, it reaches 93.8\% on eight-class colorectal cancer histology, significantly above CLBP (92.8\%). A fixed LTP coding in the same pipeline matches the learned geometry on average. We trace most of its advantage over our SM-LBP code to its absolute dead zone (Sections~\ref{sec:exp:baselines} and~\ref{sec:exp:kather}).
  \item Against the frozen ResNet-50 features (23.5 million parameters), the learned geometry has the higher mean accuracy over the six main benchmarks with 1-NN (83.10\% against 80.85\%). With LDA, the two means are similar (87.44\% against 86.74\%), and the fully learned descriptor has the highest mean (89.07\%), but both leads rest on Parquet. The ResNet-50 features are more accurate on KTH-TIPS2, USPtex and, with LDA, NewBarkTex. Without input remedies, they are also more accurate under the photometric shifts; with the same remedies, the best remedied FLEX-LBP descriptor is significantly ahead on both suites (Section~\ref{sec:exp:cnn}).
  \item We show that the learned radii, angles and thresholds remain explicit, bounded or regularized parameters whose direction of adaptation recurs across seeds. We trace the failures under noise and illuminant change to the input representation, with input-side remedies whose cost elsewhere we measure (Sections~\ref{sec:exp:photometric} and~\ref{sec:analysis}).
\end{enumerate}
The code is available at \codeurl.

\section{Related work}
\label{sec:related}

\subsection{Classical LBP and its hand-designed extensions}
\label{sec:related:classical}

CLBP~\citep{guo2010clbp} decomposes local differences into sign, magnitude and center components and shows that magnitude complements sign.
LTP~\citep{tan2010ltp} replace the zero threshold by a hand-tuned symmetric dead zone, which yields a three-valued code.
RLBP~\citep{chen2013rlbp} flips unreliable bits, and MRELBP~\citep{liu2016mrelbp} replaces point sampling with local median patches.
LBP variance~\citep{guo2010lbpv}, dominant LBP~\citep{liao2009dlbp}, cross-scale co-occurrence LBP (CS-CoLBP)~\citep{xiao2024cscolbp}, regional gradient patterns (RGP)~\citep{wang2025rgp} and sparse LBP for event-sensor data~\citep{fardo2024sparselbp} widen the design space further.
Classical descriptors, LBP included, remain competitive when data is scarce or interpretability is required~\citep{liu2019bow}.
Recent variants adapt scale and sampling through hand-crafted local statistics (EMSABP~\citep{gupta2026emsabp}, SPALBP~\citep{hu2024spalbp}, SARLBP~\citep{upadhyay2025sarlbp}).
All these methods fix their adaptation rule at design time, and the adaptive ones re-estimate their sampling per pixel.
FLEX-LBP instead learns one global set of radius, angle, threshold and color parameters from labeled data (Supplementary Section~S8 compares these design dimensions).

\subsection{Learned binary features and deep texture representations}
\label{sec:related:learned}

BSIF~\citep{kannala2012bsif} binarizes the responses of generic linear filters learned from natural images by independent component analysis (ICA), not from labeled texture data.
Local Binary Convolutional Neural Networks~\citep{juefei2017lbcnn} and LBPNet~\citep{jin2018lbpnet} build binary comparisons into deep networks, and LBPNet learns its sampling positions end to end for character and object recognition.
NLBP~\citep{peeples2024nehd} combines a sigmoid-relaxed threshold, learned $1{\times}1$ bit-weight combinations and a soft histogram layer~\citep{peeples2022histogram}.
These relaxations stay active at inference, so its deployed feature is no longer an LBP.
PCANet~\citep{chan2015pcanet}, a learned counterpart of LBP coding, learns two stages of convolution filters by principal component analysis (PCA) without labels and pools their binarized responses into histograms.
FLEX-LBP differs from these methods in two respects.
First, straight-through estimation keeps its forward pass exactly binary.
Second, it learns only a few bounded or regularized descriptor parameters per branch, each of which can be inspected.

Convolutional neural network (CNN) representations dominate texture recognition when training data is plentiful.
Generic convolutional activations transfer well to texture~\citep{cimpoi2016deep,andrearczyk2016cnn}, Deep TEN~\citep{zhang2017deepten} adds a learned residual encoding and bilinear CNNs~\citep{lin2015bilinear} model second-order interactions.
Many applied tasks, however, offer only small labeled sets, as in virus texture recognition~\citep{kylberg2011virus}, and this is the regime we target.
Training-free encodings such as RADAM~\citep{scabini2023radam} reach about 90--91\% on color Outex~TC13, but they rely on large pretrained ConvNeXt backbones and expose no descriptor parameters that can be inspected.
We do not aim to replace large-scale deep models; we ask how far learning can improve an interpretable LBP that needs no backbone.

\subsection{Training objectives and histogram postprocessing}
\label{sec:related:postproc}

Because evaluation uses nearest-neighbor matching, the training objective must shape the descriptor space for retrieval.
Neighborhood Components Analysis (NCA)~\citep{goldberger2005nca} optimizes a soft nearest-neighbor objective, Proxy-NCA~\citep{movshovitz2017proxynca} introduces scalable class proxies and additive-margin cosine losses~\citep{wang2018cosface} add a margin to the class score.
For postprocessing, power normalization was introduced for Fisher vectors~\citep{perronnin2010fisher} and also benefits compact codes~\citep{jegou2012aggregating}.

\subsection{Color and multi-scale texture descriptors}
\label{sec:related:color}

Most LBP work treats the mapping from color to gray as fixed preprocessing, although no single color space dominates across datasets~\citep{porebski2014multi}.
Other descriptors keep several color spaces and select among the resulting features~\citep{alimoussa2019compact}, as the compact hybrid multi-color space (CHMCS) descriptor~\citep{alimoussa2022chmcs} does across LBP and co-occurrence families.
Color information is strongly material-dependent~\citep{burghouts2009material}, which supports comparisons between color channels as well as within them.
Adding spatial interactions between channels to those within channels generally improves texture classification, in color and even more in hyperspectral images~\citep{porebski2022color}.
OCLBP~\citep{maenpaa2004classification} keeps one ordering of each inter-channel pair because opposing pairs such as R--G and G--R are highly redundant.
Improved OCLBP~\citep{bianconi2018ioclbp} thresholds against the local average instead of the center.
Multi-scale concatenation of two or three fixed configurations of $K$ neighbors at radius $R$ outperforms any single scale~\citep{ojala2002}.
FLEX-LBP combines these ideas: each branch learns a linear color transform, compares all nine intra- and inter-channel pairs and learns its scale within an ordered, bounded range.

\subsection{Texture analysis in histopathology}
\label{sec:related:histo}

In histopathology, the hematoxylin and eosin (H\&E) stain makes tissue classes differ in color as well as in structure.
\citet{kather2016multi} released a colorectal cancer texture benchmark and reached 87.4\% with combined gray-level descriptors, LBP included, and a support vector machine.
Stain color varies between laboratories, scanners and staining batches, which motivates stain normalization~\citep{reinhard2001color,macenko2009normalizing} and color augmentation~\citep{tellez2019quantifying}.
Whether color normalization helps depends on whether training and test images differ in color.

\section{Method}
\label{sec:method}

We first recall the classical LBP and state what FLEX-LBP learns (Section~\ref{sec:method:background}).
We then describe the architecture, the operator and the parameters of each learned component (Sections~\ref{sec:method:arch}--\ref{sec:method:geometry}).
Finally, we explain how gradients cross the binary steps (Section~\ref{sec:method:ste}) and state the training objective (Section~\ref{sec:method:loss}) and the postprocessing and classification (Section~\ref{sec:method:postprocessing}).
We write descriptor and parameter vectors in bold and matrices in upper case.

\subsection{From the classical LBP to FLEX-LBP}
\label{sec:method:background}

The classical LBP~\citep{ojala2002} describes a texture by how each pixel compares with its surroundings.
Around each pixel $p$ of a gray-level image $x$, it samples $K$ neighbors on a circle of radius $R$, at the evenly spaced angles $\theta_k = 2\pi k/K$.
Each neighbor yields one bit: 1 if it is at least as bright as the center pixel and 0 otherwise.
The $K$ bits of a pixel are packed into one integer code,
\begin{equation}
  \mathrm{LBP}_{K,R}(p) = \sum_{k=0}^{K-1} 2^k\, \mathbf{1}\!\left[x(p + \Delta_k) - x(p) \geq 0\right],
  \label{eq:lbp}
\end{equation}
where $\Delta_k = R\,(\cos\theta_k, \sin\theta_k)$ is the offset of the $k$-th neighbor and $\mathbf{1}[\cdot]$ equals 1 if its condition holds and 0 otherwise.
Neighbors at non-integer positions are read by bilinear interpolation.
With $K{=}8$, each pixel receives a code between 0 and 255 that names its local pattern, such as an edge or a corner.
The descriptor of the image is the 256-bin histogram of these codes over all its pixels.

\begin{figure*}[tbp]
\centering
\includegraphics[width=\textwidth]{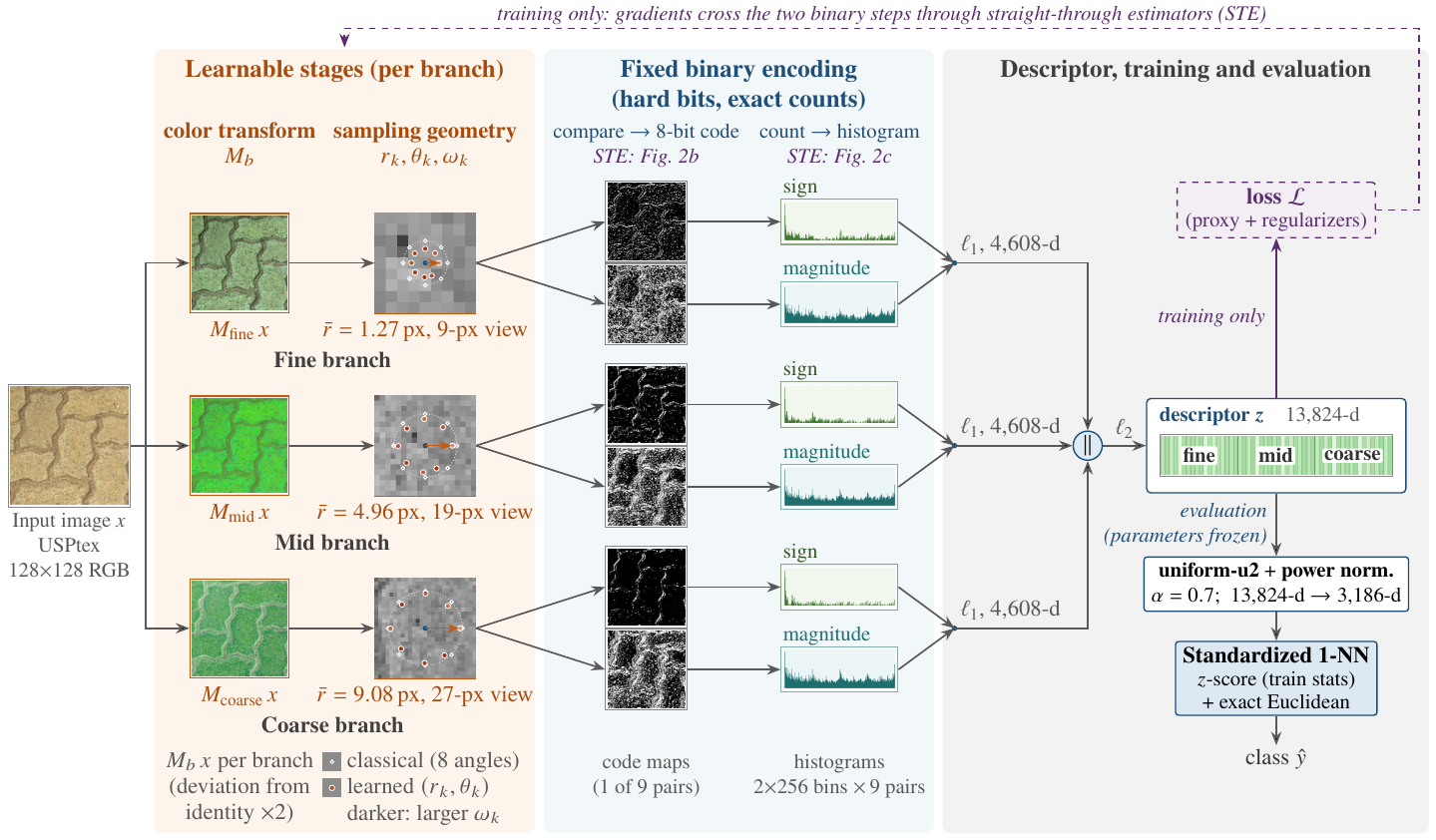}
\caption{Overview of FLEX-LBP on one real USPtex texture, with the parameters learned on USPtex by the All joint configuration (Section~\ref{sec:exp:ablation}; symbols in Sections~\ref{sec:method:arch}--\ref{sec:method:geometry}, with the branch index $b$ dropped). Left (learnable stages): each branch applies its learned color transform $M_b$ and samples $K{=}8$ neighbors at learned per-direction radii $r_k$ and angles $\theta_k$, with mean radius $\bar r$. Open circles mark the classical initialization; filled dots mark the learned positions, shaded by the learned threshold $\omega_k$ relative to the largest threshold of the branch. The geometry panels show the central 9, 19 and 27 pixels of the gray-level input. Middle (fixed binary encoding): hard comparisons give sign and magnitude codes, counted exactly into 256-bin histograms for each channel pair. Right: the histograms are normalized and concatenated into the $13{,}824$-dimensional descriptor $\boldsymbol{z}$ (Eq.~\eqref{eq:descriptor}). During training, the loss (proxy loss and regularizers, Eq.~\eqref{eq:loss}) sends gradients back to every learnable stage (dashed); straight-through estimators carry them across the two binary steps (Fig.~\ref{fig:ste}b,c). For evaluation, with frozen parameters, $\boldsymbol{z}$ is projected onto uniform patterns, power-normalized ($3{,}186$ dimensions) and classified by standardized one-nearest-neighbor (1-NN) matching against the training images. The ablation of Table~\ref{tab:ablation} skips this postprocessing. The color-transform panels show $M_b x$ with its deviation from the identity doubled for visibility. Input image from USPtex~\citep{backes2012color}.}
\label{fig:joint}
\end{figure*}

This operator leaves five choices to its designer: the radius $R$, the sampling angles $\theta_k$, the comparison threshold (zero in Eq.~\eqref{eq:lbp}), the input color space and the number of gray levels.
FLEX-LBP keeps the stages of this operator and learns these five choices by backpropagation.
It has three branches, each with 30 learnable parameters, 90 in total.
A branch stores nine values of a color transform, four radii (the sampling distance can differ mildly along four axes), eight angles, eight thresholds and one number of gray levels.
The experiments compare configurations that learn subsets of these parameters (Section~\ref{sec:exp:ablation}).
FLEX-LBP also adopts two established extensions of the operator: the sign-plus-magnitude coding of CLBP~\citep{guo2010clbp} and the comparisons between color channels of OCLBP~\citep{maenpaa2004classification}.
Because the forward pass still computes hard comparisons and exact histogram counts, inference costs the same as a fixed multi-scale color LBP with the same sampling (timings in Section~\ref{sec:exp:details}).
Fig.~\ref{fig:joint} summarizes the pipeline on one real USPtex texture.

\subsection{Multi-branch architecture}
\label{sec:method:arch}

FLEX-LBP has three parallel branches that sample the input image at different radii, and thus at different spatial scales: fine, mid and coarse.
Each branch $b \in \{\mathrm{fine}, \mathrm{mid}, \mathrm{coarse}\}$ applies its own learnable color transform and LBP operator to the RGB input image $x \in \mathbb{R}^{3 \times H \times W}$.
It produces a sign-plus-magnitude histogram descriptor, the SM-LBP descriptor
\begin{equation}
  \boldsymbol{h}_b = \mathrm{SMLBP}\!\left(M_b\, x;\; \boldsymbol{r}_b, \boldsymbol{\theta}_b, \boldsymbol{\omega}_b, L_b\right),
  \label{eq:branch}
\end{equation}
where $M_b \in \mathbb{R}^{3 \times 3}$ is the learnable color transform, $\boldsymbol{r}_b$ and $\boldsymbol{\theta}_b$ are the radii and angles of the $K$ sampling directions, $\boldsymbol{\omega}_b$ holds the per-direction thresholds and $L_b$ is the number of quantization levels.
Sections~\ref{sec:method:smlbp}--\ref{sec:method:geometry} define them and show how learnable parameters set the radii and angles.
The branch descriptors are then normalized and concatenated into the global descriptor
\begin{equation}
  \boldsymbol{z} = \ell_2\text{-}\mathrm{norm}\!\left[\ell_1\text{-}\mathrm{norm}(\boldsymbol{h}_{\mathrm{fine}});\; \ell_1\text{-}\mathrm{norm}(\boldsymbol{h}_{\mathrm{mid}});\; \ell_1\text{-}\mathrm{norm}(\boldsymbol{h}_{\mathrm{coarse}})\right],
  \label{eq:descriptor}
\end{equation}
that is, each branch histogram is $\ell_1$-normalized and the concatenation is $\ell_2$-normalized once.

\subsection{Sign-plus-magnitude LBP operator}
\label{sec:method:smlbp}

The SM-LBP operator extends Eq.~\eqref{eq:lbp} in two ways.
First, following the CLBP decomposition~\citep{guo2010clbp}, it produces two codes instead of one: one encodes the sign of the center--neighbor difference and the other its magnitude.
The sign pattern captures local ordinal structure such as edges and corners, while the magnitude pattern records the strength of local contrast, which is discriminative on material and bark textures~\citep{guo2010clbp,liu2019bow}.

Second, it compares the channels of a color image.
Each branch evaluates all $3 \times 3 = 9$ channel pairs $(i, j)$ of its transformed image $\tilde{x}_b = M_b\, x$, center channel $i$ against neighbor channel $j$.
The three intra-channel pairs ($i{=}j$) capture the usual gray-level texture, and the six inter-channel pairs ($i{\neq}j$) the contrast between color channels.
Comparing a center in one channel with neighbors in another follows OCLBP~\citep{maenpaa2004classification}, which keeps only one ordering of each inter-channel pair.
We keep both, so that every channel serves as the center for every other one.

For branch $b$, channel pair $(i, j)$, direction $k \in \{0, \ldots, K{-}1\}$ and pixel $p$, the thresholded difference is
\begin{equation}
  d_{b,k}^{(i,j)}(p) = \tilde{x}_{b,j}\!\left(p + \Delta_{b,k}\right) - \tilde{x}_{b,i}(p) - \omega_{b,k},
  \label{eq:signed_diff}
\end{equation}
where $\Delta_{b,k} = r_{b,k}\,(\cos\theta_{b,k}, \sin\theta_{b,k})$ is the sub-pixel offset of the $k$-th neighbor, now with its own learned radius and angle.
The learned threshold $\omega_{b,k}$ starts at zero, its value in Eq.~\eqref{eq:lbp}.
Bilinear interpolation at the non-integer position $p + \Delta_{b,k}$ makes the difference differentiable with respect to $r_{b,k}$ and $\theta_{b,k}$.

The sign bit is $\beta_{b,k}^{(i,j)}(p) = \mathbf{1}\!\left[d_{b,k}^{(i,j)}(p) > 0\right]$, with ties counted as 0, and the $K$ sign bits are packed into the integer code $\sum_{k=0}^{K-1} 2^k \beta_{b,k}^{(i,j)}(p) \in \{0, \ldots, 2^K{-}1\}$.
The magnitude is the threshold-free contrast $m_{b,k}^{(i,j)}(p) = \bigl|\tilde{x}_{b,j}(p + \Delta_{b,k}) - \tilde{x}_{b,i}(p)\bigr|$.
Following the $\mathrm{CLBP\_M}$ construction~\citep{guo2010clbp}, each magnitude is binarized against the mean magnitude of the current image, $\mu_{b}^{(i,j)} = \mathrm{mean}_{k,p}\,m_{b,k}^{(i,j)}(p)$.
The $K$ magnitude bits $\mathbf{1}\!\left[m_{b,k}^{(i,j)}(p) > \mu_{b}^{(i,j)}\right]$ are packed into a second code in the same way.
Both codes therefore take $2^K$ values, so the uniform-pattern projection of Section~\ref{sec:method:postprocessing} applies identically to both histograms.
Because $\mu_{b}^{(i,j)}$ scales with the contrast of the image, the magnitude code is unchanged when all intensities are multiplied by a common factor.
It does not, however, record the absolute size of the differences, which for an inter-channel pair also reflects the color of the texture; Section~\ref{sec:exp:baselines} measures what this costs and saves.

Each channel pair thus yields one sign histogram and one magnitude histogram.
With $K{=}8$, a branch descriptor has $9 \times (256 + 256) = 4{,}608$ bins, and the three-branch descriptor has $13{,}824$ bins before postprocessing.
Supplementary Fig.~S2 works through one sign code and one magnitude code at a real pixel.

\subsection{Learnable color transform and quantization}
\label{sec:method:color}

Because no single color space is best across datasets~\citep{porebski2014multi}, each branch learns its own linear color basis, a $3 \times 3$ matrix $M_b$ initialized near the identity,
\begin{equation}
  \tilde{x}_b = M_b\, x, \qquad M_b^{(0)} = I_3 + \epsilon\, E, \quad \epsilon = 0.01,
  \label{eq:color}
\end{equation}
where $E$ has independent standard normal entries.
Each channel of $\tilde{x}_b$ is thus a weighted sum of R, G and B, and training starts almost from the RGB channels themselves.

Before the comparisons of Eq.~\eqref{eq:signed_diff}, each branch clips $\tilde{x}_b$ to $[0,1]$ and rounds it to $L_b$ uniform gray levels, $\mathrm{round}\big((L_b{-}1)\,\tilde{x}_b\big)/(L_b{-}1)$, with $L_b \in [8, 256]$ bounded by a sigmoid.
The number of levels is learned only in the Mixer+Q configuration (Section~\ref{sec:exp:ablation}), which initializes it at 128; all other configurations freeze it at 256, the resolution of 8-bit images.

\subsection{Bounded geometry parameterization}
\label{sec:method:geometry}

The radii and angles are not learned directly: each is a bounded function of unconstrained parameters, which keeps the learned configuration interpretable and prevents degenerate solutions.

\paragraph{Radius}
A branch radius is a function of an unconstrained learnable scalar $u_b$, bounded by the logistic sigmoid $\sigma$,
\begin{equation}
  r_b = r_b^{\min} + \left(r_b^{\max} - r_b^{\min}\right) \sigma(u_b),
  \label{eq:radius}
\end{equation}
so it stays within a scale-specific range.
The three ranges are ordered (fine $<$ mid $<$ coarse) and mostly disjoint, which discourages two branches from converging to the same radius.
The bounds and initial radii depend on the dataset (Section~\ref{sec:exp:details} and Supplementary Section~S4).

A single radius would keep all neighbors on a circle.
To let the neighborhood stretch mildly in some directions (anisotropy), a branch learns four such basis radii $r_b^{(a)}$, $a = 1, \ldots, 4$, each bounded as in Eq.~\eqref{eq:radius}.
Each basis radius is attached to a fixed axis through the center pixel, at the angle $\phi_a = (a{-}1)\pi/4$ (0, 45, 90 or 135 degrees).
Each sampling direction takes a soft average of the basis radii, with the largest weights on the axes closest to it,
\begin{equation}
  r_{b,k} = \sum_{a=1}^{4} w_{k,a}\, r_b^{(a)}, \qquad w_{k,a} = \mathrm{softmax}_a\!\big(\kappa \cos 2(\theta_{b,k} - \phi_a)\big),
  \label{eq:directional}
\end{equation}
with a fixed sharpness $\kappa{=}8$, so the weights add no parameters and follow the learned angles.
As a convex combination of these bounded radii, each $r_{b,k}$ stays within the range of its branch, and the anisotropy is bounded by the spread of the basis radii.
The factor 2 in the cosine assigns opposite directions the same weights, so at the initial angles each axis serves two opposite neighbors and the eight neighbors share the four radii.
All four start at the initial radius of the branch, so the neighborhood starts as a circle.

\paragraph{Angles}
The $K$ sampling directions are cumulative sums of positive increments,
\begin{equation}
  \theta_{b,k} = 2\pi \sum_{k' < k} \delta_{b,k'}, \qquad \boldsymbol{\delta}_b = \mathrm{softmax}(\boldsymbol{\gamma}_b),
  \label{eq:angles}
\end{equation}
so $\theta_{b,0}{=}0$, and $\boldsymbol{\gamma}_b \in \mathbb{R}^K$ is unconstrained and learnable (initialized at zero, i.e., uniform spacing).
The softmax makes the increments positive and makes them sum to one.
The directions are therefore always ordered and cover the full circle, while their spacing can become non-uniform; the angular-gap penalty of $\mathcal{L}_{\mathrm{geom}}$ (Section~\ref{sec:method:loss}) discourages collapse.
Adding the same constant to all values of $\boldsymbol{\gamma}_b$ leaves the softmax unchanged, so these eight parameters carry seven degrees of freedom.
The first direction stays at angle zero, and the other seven can move.

\subsection{Differentiability: straight-through estimator}
\label{sec:method:ste}

Training needs the derivative of the loss with respect to every descriptor parameter.
Two steps of the SM-LBP pipeline, however, are piecewise constant: the comparison that produces a bit and the counting that turns codes into a histogram.
Their derivatives are zero almost everywhere, so plain backpropagation of the training loss $\mathcal{L}$ (Section~\ref{sec:method:loss}) would leave every LBP parameter unchanged.
Soft bits or soft histograms in the forward pass would provide gradients, but the trained descriptor would then differ from the deployed one and would no longer be an LBP.

\begin{figure*}[tbp]
\centering
\includegraphics{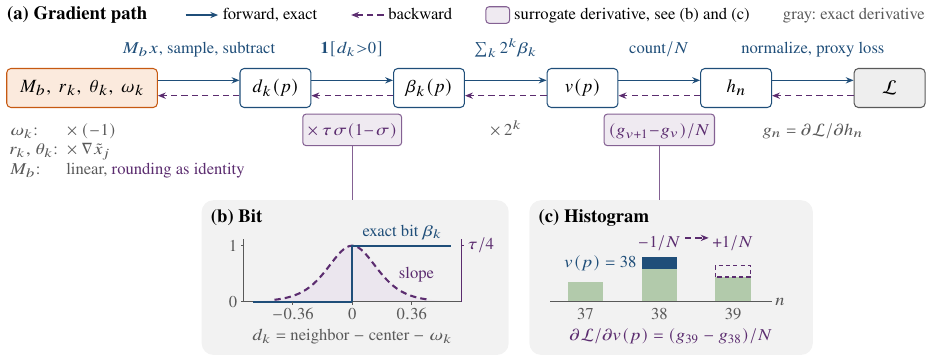}%
\caption{Straight-through estimator (STE) of FLEX-LBP for one sign comparison. (a)~Gradient path from the learnable parameters (orange, as in Fig.~\ref{fig:joint}) to the loss. The forward pass (navy, solid) is exact; the backward pass (purple, dashed) multiplies exact derivatives (gray) by two surrogate derivatives (purple boxes), detailed in (b) and (c). Labels with $\times$ are local factors, and the other two are the gradients that arrive at $h_n$ and $v(p)$, so their product from $v(p)$ to $d_k(p)$ is Eq.~\eqref{eq:ste_chain}. Under the parameter node, $\omega_k$ receives the factor $-1$, and $r_k$ and $\theta_k$ receive the radial and tangential changes of the image through its gradient $\nabla\tilde{x}_j$. The color transform $M_b$ acts linearly, and the rounding to $L_b$ gray levels passes its gradient unchanged (purple). (b)~The bit $\beta_k$ is an exact step in $d_k$ (navy). The backward pass uses the slope $\tau\sigma(1-\sigma)$ of the tempered sigmoid $\sigma = \sigma(\tau d_k)$, $\tau{=}10$, drawn with its peak $\tau/4$ at height 1; it falls below a tenth of its peak for $|d_k| > 0.36$. (c)~A pixel with code $v(p) = 38$ adds $1/N$ to bin 38 (navy). In the backward pass, linear interpolation moves this vote toward bin 39 at a rate of $1/N$ per unit of $v(p)$ (purple), which yields Eq.~\eqref{eq:ste_hist}.}
\label{fig:ste}
\end{figure*}

We therefore use the STE~\citep{bengio2013ste,hubara2016bnn}, which keeps the forward pass exact and changes only the backward pass.
For a piecewise-constant operation $q(u)$, we choose a smooth approximation $\tilde{q}(u)$, its surrogate, and the network computes
\begin{equation}
  q_{\mathrm{STE}}(u) = q(u) + \tilde{q}(u) - \mathrm{sg}\!\left[\tilde{q}(u)\right],
  \label{eq:ste}
\end{equation}
where the stop-gradient operator $\mathrm{sg}[\cdot]$ returns its argument in the forward pass and has zero derivative.
The last two terms cancel in the forward pass, so $q_{\mathrm{STE}}(u) = q(u)$ exactly, while backpropagation uses $\partial q_{\mathrm{STE}}/\partial u = \tilde{q}\,'(u)$.
In other words, the network computes with the exact step and differentiates as if it were the smooth surrogate.
FLEX-LBP applies Eq.~\eqref{eq:ste} to the bits and to the histogram (Fig.~\ref{fig:ste}); below, we omit the indices $b$, $(i,j)$ and $p$ when they are clear.

\paragraph{Bits}
The surrogate of the sign bit $\beta_k = \mathbf{1}[d_k > 0]$ of Eq.~\eqref{eq:signed_diff} is the tempered sigmoid $\sigma(\tau d_k)$, a smooth step that steepens as the temperature $\tau$ grows ($\tau{=}10$ here), so that
\begin{equation}
  \frac{\partial \beta_k}{\partial d_k} \approx \tau\,\sigma(\tau d_k)\bigl(1 - \sigma(\tau d_k)\bigr).
  \label{eq:ste_bit}
\end{equation}
This slope peaks at $\tau/4$ where the difference equals the threshold and falls below a tenth of its peak for $|d_k| > 0.36$.
With intensities in $[0,1]$, the gradient therefore comes from comparisons close to their decision boundary.
We keep $\tau{=}10$ in all experiments; Supplementary Section~S16 reports $\tau{=}3$ and $\tau{=}30$.
The magnitude bits $\mathbf{1}[m_k > \mu]$ use the same surrogate on $m_k - \mu$.
The image mean $\mu$ is recomputed in every forward pass but treated as a constant in the backward pass.
Each magnitude surrogate thus stays local to its own comparison, as for the sign bits; we did not test backpropagating through $\mu$.
The integer code $v = \sum_k 2^k \beta_k$ is linear in the bits, so $\partial v/\partial \beta_k = 2^k$ holds exactly.

\paragraph{Histogram}
A histogram bin is the fraction of the $N$ pixels whose code falls into it, $h_n = \tfrac{1}{N}\sum_p \mathbf{1}[v(p) = n]$.
The surrogate treats the code as a real number and shares its vote between its two neighboring bins by linear interpolation: bin $\lfloor v \rfloor$ receives $1 - (v - \lfloor v \rfloor)$ and bin $\lfloor v \rfloor + 1$ the rest.
The forward pass produces only integer codes, where the interpolation has a kink, so we use its slope to the right of the code.
Raising $v(p)$ moves the vote of pixel $p$ from bin $v(p)$ toward bin $v(p)+1$: the upper bin gains $1/N$ per unit of $v(p)$ and the lower bin loses $1/N$, so
\begin{equation}
  \frac{\partial \mathcal{L}}{\partial v(p)} = \frac{1}{N}\left(g_{v(p)+1} - g_{v(p)}\right), \qquad g_n = \frac{\partial \mathcal{L}}{\partial h_n}.
  \label{eq:ste_hist}
\end{equation}
A code is pushed up when the loss favors one more pixel in the next bin over one more pixel in the current bin.
The top code $v(p) = 2^K - 1$ has no upper neighbor: its interpolation is clamped to the last bin, so its surrogate slope is zero.

\paragraph{From the histogram to the parameters}
Chaining Eqs.~\eqref{eq:ste_bit} and~\eqref{eq:ste_hist}, the difference $d_k(p)$ of one sign comparison receives
\begin{equation}
  \frac{\partial \mathcal{L}}{\partial d_k(p)} = \frac{2^k}{N}\left(g_{v(p)+1} - g_{v(p)}\right)\tau\,\sigma(\tau d_k(p))\bigl(1 - \sigma(\tau d_k(p))\bigr),
  \label{eq:ste_chain}
\end{equation}
and every parameter sums such terms over pixels, directions, channel pairs and histograms through ordinary derivatives.
In words, a comparison is pushed according to whether the loss prefers the next bin of its pixel to the current one.
Only comparisons close to their decision boundary receive a sizable push.
Eq.~\eqref{eq:ste_chain} extrapolates the change between bins $v$ and $v+1$ to the flip of bit $k$, which moves the pixel by $2^k$ bins, so the estimate is biased.
Section~\ref{sec:exp:significance} and Supplementary Section~S16 compare it with a bit-aware alternative that reads the histogram gradient at the two codes that the flip actually connects.

The remaining factors are ordinary derivatives: a threshold enters with $\partial d_k/\partial \omega_k = -1$.
The radius and the angle act through the bilinear interpolation of $\tilde{x}_j$ at $p + \Delta_k$: moving a sampling point changes the sampled intensity according to the spatial gradient of the image.
With $\nabla\tilde{x}_j$ this gradient at that point, $\partial d_k/\partial r_k = \nabla\tilde{x}_j \cdot (\cos\theta_k, \sin\theta_k)$ and $\partial d_k/\partial \theta_k = r_k\, \nabla\tilde{x}_j \cdot (-\sin\theta_k, \cos\theta_k)$ are the radial and tangential intensity changes.
They reach the unconstrained parameters through the parameterizations of Section~\ref{sec:method:geometry}.
Because the color transform is linear, $d_k$ depends on row $j$ of $M_b$ through the RGB value of the neighbor and on row $i$ through that of the center.
The magnitudes $m_k = |d_k + \omega_k|$ do not depend on the thresholds; they pass gradients to the geometry and the color transform with the factor $\mathrm{sign}(d_k + \omega_k)$.

\paragraph{Quantization}
The rounding of Section~\ref{sec:method:color} is also piecewise constant.
The backward pass treats it as the identity for $\tilde{x}_b$, the classical STE for rounding~\citep{bengio2013ste}.
For $L_b$, the rounded integer counts as a constant, so $L_b$ receives a gradient only through the factor $1/(L_b - 1)$.

\paragraph{What the estimator guarantees}
Because all surrogates act only in the backward pass, the descriptor computed during training is the one used at test time.
It is a hard-binary LBP at the learned $(M_b, \boldsymbol{r}_b, \boldsymbol{\theta}_b, \boldsymbol{\omega}_b, L_b)$, whose parameters can be read off.
Recomputing the bits, codes and histogram counts of Fig.~\ref{fig:joint} from the intensities sampled by the model, with integer bit packing and counting, reproduces the model output exactly for all 27 channel pairs.
The color transform and the bilinear interpolation remain floating-point operations, as in any interpolated LBP, so the exactness concerns the binary coding and counting, not the sampled intensities.
With its color transform and thresholds, FLEX-LBP is a generalized LBP rather than a classical LBP with other radii; what it shares with the classical operator is exact hard-binary inference.
Like every STE, the estimator is biased, which can make learning slower or less effective but cannot make the deployed descriptor differ from the one that was trained and evaluated.

\subsection{Training objective}
\label{sec:method:loss}

All experiments use a proxy-only descriptor-learning objective~\citep{movshovitz2017proxynca}.
Each class has one learned unit vector, its proxy, and the loss pulls the descriptor of an image toward the proxy of its class and away from the others.
All discriminative pressure therefore comes from one metric-learning loss on the $\ell_2$-normalized descriptor, without a classification head or data augmentation, which isolates the effect of learning on the descriptor.
The loss is
\begin{equation}
  \mathcal{L} = \mathcal{L}_{\mathrm{proxy}} + \lambda_{\mathrm{geom}}\,\mathcal{L}_{\mathrm{geom}} + \lambda_{\omega}\,\mathcal{L}_{\omega}.
  \label{eq:loss}
\end{equation}
$\mathcal{L}_{\mathrm{proxy}}$ is a cosine proxy-classification loss with unit-norm learnable class proxies $\boldsymbol{p}_c$, which are discarded at inference, and an additive margin $\xi$ on the true class $y$~\citep{wang2018cosface},
\begin{equation}
  \mathcal{L}_{\mathrm{proxy}} = -\log \frac{e^{s(\cos(\boldsymbol{z},\boldsymbol{p}_y) - \xi)}}{e^{s(\cos(\boldsymbol{z},\boldsymbol{p}_y) - \xi)} + \sum_{c \neq y} e^{s\cos(\boldsymbol{z},\boldsymbol{p}_c)}},
  \label{eq:proxy}
\end{equation}
where $s$ is a scale.
$\mathcal{L}_{\mathrm{geom}}$ keeps the geometry physically plausible: it combines a quadratic pull of the radii toward their initialization, a penalty on angular gaps below 0.3\,rad and a mild pull of the angles toward the standard LBP directions, with relative weights 1, 0.5 and 0.3.
$\mathcal{L}_{\omega}$ is an $\ell_2$ penalty on the thresholds $\omega_{b,k}$ that prevents unbounded threshold growth.
Each regularizer is active only when the parameters it constrains are learned, and Section~\ref{sec:exp:details} lists the values of $s$, $\xi$, $\lambda_{\mathrm{geom}}$ and $\lambda_{\omega}$.

\subsection{Descriptor postprocessing and classification}
\label{sec:method:postprocessing}

Before classification, two transformations are applied to the histograms, identically for fixed and learned descriptors.

\paragraph{Uniform-pattern projection}
Following \citet{ojala2002}, each 256-bin histogram block is projected onto the uniform patterns (denoted u2), the codes with at most two bitwise transitions around the circle.
For $K{=}8$ this reduces each block from 256 to 59 bins: 58 uniform patterns and one bin that pools all non-uniform codes, which merges the noisy tail of rare patterns.

\paragraph{Power normalization}
Element-wise power normalization~\citep{perronnin2010fisher} is then applied to the projected descriptor $\boldsymbol{z}$,
\begin{equation}
  \hat{z}_n = \mathrm{sign}(z_n)\,|z_n|^{\alpha}, \qquad \hat{\boldsymbol{z}} \leftarrow \hat{\boldsymbol{z}} / \|\hat{\boldsymbol{z}}\|_2,
  \label{eq:powernorm}
\end{equation}
with $\alpha{=}0.7$, which compresses the dynamic range of the bins and up-weights rare patterns.
The postprocessed descriptor has $3 \times 9 \times 2 \times 59 = 3{,}186$ dimensions.

\paragraph{Classification}
Descriptors are $z$-score standardized with training-set statistics and classified by exact Euclidean nearest-neighbor matching, which we call standardized 1-NN.
This non-parametric readout measures the geometry of the descriptor space without any learned classifier and is the primary metric throughout the paper.
As a second readout, we fit an LDA on the standardized training descriptors, with the covariance shrinkage set automatically by the Ledoit--Wolf estimator~\citep{ledoit2004shrinkage}.
LDA has no hyperparameter to tune but, unlike 1-NN, is a classifier fitted to the labeled training data.

\section{Experiments}
\label{sec:experiments}

\subsection{Datasets}
\label{sec:exp:datasets}

\begin{figure*}[tbp]
\centering
\includegraphics[width=\textwidth]{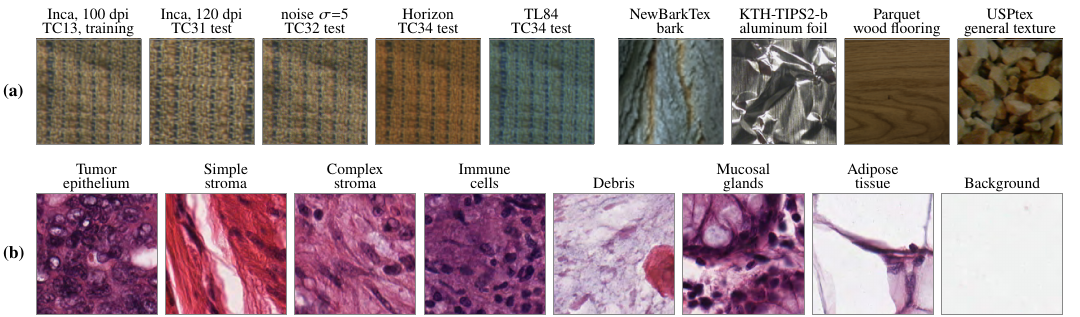}
\caption{Sample images. (a)~Left: one Outex texture (canvas001, first crop) under the training conditions of TC13, TC31, TC32 and TC34 (Inca illuminant, 100\,dpi). The next four tiles show it under the test conditions of TC31 (120\,dpi), TC32 (the same image with Gaussian noise, $\sigma{=}5$) and TC34 (Horizon and TL84 illuminants). Right: one image each of NewBarkTex, KTH-TIPS2-b, Parquet and USPtex, chosen for legibility. All are shown at the $128{\times}128$ input size (Section~\ref{sec:exp:details}). The noise is hardly visible, yet $\mathrm{LBP}^{riu2}$ reaches only 13.24\% on TC32 (Table~\ref{tab:ablation}). (b)~The eight tissue classes of Kather-CRC: the first tile of each class in file order ($150{\times}150$ pixels, $74{\times}74$\,\textmu m, H\&E stain). Images from Outex~\citep{ojala2002outex,lille2022extendedoutex}, NewBarkTex~\citep{porebski2014newbarktex}, KTH-TIPS2-b~\citep{mallikarjuna2006kthtips}, Parquet~\citep{bianconi2013parquet}, USPtex~\citep{backes2012color} and Kather-CRC~\citep{kather2016multi} (CC BY 4.0).}
\label{fig:samples}
\end{figure*}

We evaluate on nine benchmarks (Table~\ref{tab:datasets}, Fig.~\ref{fig:samples} and Supplementary Section~S19).
Six texture benchmarks of bark, materials, wood and general textures form the main evaluation, and two Outex suites with photometric shifts are analyzed separately.
A colorectal cancer (CRC) histology benchmark, Kather-CRC, serves as an application.

\begin{table}[tbp]
\centering
\small
\caption{The nine benchmarks, all evaluated in RGB; the split lists the numbers of training and test images. The Outex benchmarks are 68-class color suites: the original Outex\_TC\_00013 and the extended Outex\_TC\_00031, 00032 and 00034. Kather-CRC is evaluated by stratified five-fold cross-validation at the tile level.}
\label{tab:datasets}
\renewcommand{\arraystretch}{1.1}
\setlength{\tabcolsep}{4pt}
\begin{adjustbox}{max width=\linewidth}
\begin{tabular}{@{}lrrl@{}}
\toprule
\textbf{Dataset} & \textbf{Classes} & \textbf{Images} & \textbf{Split (train/test)} \\
\midrule
\multicolumn{4}{@{}l}{\textit{Main benchmarks}} \\
NewBarkTex~\citep{porebski2014newbarktex} & 6 & 1,632 & Official, 816/816 \\
KTH-TIPS2-b~\citep{caputo2005class,mallikarjuna2006kthtips}  & 11  & 4,749 & Three samples / sample~a \\
Outex~TC13~\citep{ojala2002outex}    & 68  & 1,360 & Official, 680/680 \\
Outex~TC31~\citep{lille2022extendedoutex} & 68  & 2,720 & Official, 1,360/1,360 \\
Parquet~\citep{bianconi2013parquet}       & 38  & 228   & First/last three per class \\
USPtex~\citep{backes2012color}       & 191 & 2,292 & Standard, 1,146/1,146 \\
\midrule
\multicolumn{4}{@{}l}{\textit{Photometric shift suites}} \\
Outex~TC32~\citep{lille2022extendedoutex} & 68  & 2,720 & Official, 1,360/1,360 \\
Outex~TC34~\citep{lille2022extendedoutex} & 68  & 4,080 & Official, 1,360/2,720 \\
\midrule
\multicolumn{4}{@{}l}{\textit{Application}} \\
Kather-CRC~\citep{kather2016multi} & 8 & 5,000 & Five folds, 4,000/1,000 each \\
\bottomrule
\end{tabular}
\end{adjustbox}
\end{table}

\textbf{NewBarkTex}~\citep{porebski2014newbarktex} (six bark classes) uses its official split, and \textbf{USPtex}~\citep{backes2012color} (191 classes of general textures) the standard decomposition distributed by the LISIC laboratory.
\textbf{KTH-TIPS2-b}~\citep{caputo2005class,mallikarjuna2006kthtips}, KTH-TIPS2 hereafter, holds out physical sample~a of each of its 11 materials as the test set and trains on the other three samples.
This is the easier of the two common KTH-TIPS2 conventions, so published results that train on a single sample are not directly comparable to ours.
\textbf{Parquet}~\citep{bianconi2013parquet} holds 38 classes of wood flooring (14 species in 2--4 visually similar grades), with six images per class.
We train on the first three images of each class and test on the last three.
This fixed split differs from the alternating split of the public LISIC decomposition, so our Parquet results are not directly comparable to published ones.
\textbf{Outex}: in the original color suite TC13~\citep{ojala2002outex}, the training and test images are acquired under identical conditions.
In the extended Outex suites TC31, TC32 and TC34~\citep{lille2022extendedoutex}, which share their training images (Inca illuminant, 100\,dpi), the test images differ in one controlled factor.
This factor is the spatial resolution in TC31 (120\,dpi), additive Gaussian noise ($\sigma{=}5$) in TC32 and the illuminant in TC34 (Horizon and TL84).
Fig.~\ref{fig:samples}a shows one texture under the four conditions.

TC32 and TC34 change the photometry of the test images rather than their texture.
Our color descriptor is invariant to neither change by design.
We therefore analyze these two suites separately (Section~\ref{sec:exp:photometric}), report them in every table and exclude them from the averages over the main benchmarks.
This grouping was decided after their results and their diagnosis were known.

\textbf{Kather-CRC}~\citep{kather2016multi}, named after its first author and colorectal cancer, contains 5,000 RGB tiles of $150{\times}150$ pixels cut from ten H\&E-stained slides, 625 for each of eight tissue classes (Fig.~\ref{fig:samples}b).
Our five folds, like the ten of \citet{kather2016multi}, are formed at the tile level, so tiles of one slide and of neighboring regions fall into both training and test folds.
The benchmark therefore measures recognition within ten slides from one center, not generalization to new patients.

\subsection{Ablation design}
\label{sec:exp:ablation}

All configurations share the architecture, the proxy-only loss of Eq.~\eqref{eq:loss}, the optimizer and the evaluation protocol (standardized 1-NN, Section~\ref{sec:method:postprocessing}).
They differ only in which descriptor parameters are learned, except that Mixer+Q starts its quantization at 128 levels instead of 256:
\begin{itemize}
  \item \emph{Fixed} learns no descriptor parameter: it keeps the initial radii, the uniform angles, the zero thresholds and $M_b^{(0)}$, trains only the class proxies and is the no-learning reference;
  \item \emph{Color} learns the color transforms $M_b$ only;
  \item \emph{Radius} learns the four basis radii only;
  \item \emph{Threshold} learns the thresholds $\omega_{b,k}$ only;
  \item \emph{Geometry} learns the radii, angles and thresholds jointly, with $M_b$ and $L_b$ frozen;
  \item \emph{Mixer+Q} learns the color transforms $M_b$, which mix the channels, and the numbers of quantization levels $L_b$ jointly, with the radii, angles and thresholds frozen;
  \item \emph{All joint}, the fully learned descriptor, learns the color transforms, radii, angles and thresholds jointly, with the quantization frozen.
\end{itemize}
Training lasts 8 to 20 epochs depending on the dataset, and we select the checkpoint by validation 1-NN accuracy on a 15\% stratified holdout of the training partition.
On KTH-TIPS2, an inner leave-one-sample-out validation chooses the number of epochs instead (Supplementary Section~S1).
The test partition is never used for checkpoint selection.

Every configuration is trained with seed 42; Geometry is also trained with seeds 43--46 and All joint with seeds 43 and 44.
The seed sets the batch order, the validation holdout and the random initialization $M_b^{(0)}$ of Eq.~\eqref{eq:color}.
The Fixed descriptor also keeps this initialization, so we compare each learned run with the Fixed descriptor of its own seed.

\subsection{Implementation details}
\label{sec:exp:details}

\paragraph{Architecture}
All images are resized to $128{\times}128$ pixels.
The radius bounds and initial radii follow the ordered fine/mid/coarse scheme of Section~\ref{sec:method:geometry} in two templates (Supplementary Section~S4 lists the exact bounds).
The compact one serves NewBarkTex and Parquet (initial radii of 2.0, 5.0 and 8.0 pixels), and the wide one the other benchmarks (2.0, 6.0 and 10.0 pixels).
Within a dataset, every configuration shares the same template.
The templates were set during development, when the test accuracies of earlier runs were visible, and not by a held-out selection procedure, so Supplementary Section~S16 swaps them on every benchmark.

\paragraph{Training}
We use Adam with component-specific learning rates: $2{\times}10^{-3}$ for radii; $10^{-3}$ for color transforms, angles and class proxies; and $5{\times}10^{-4}$ for thresholds and quantization.
Weight decay is $10^{-4}$, except on radii, angles, thresholds and quantization, and a step schedule halves all rates once.
The batch size is 4 on NewBarkTex and Parquet and 8 elsewhere.
The loss of Eq.~\eqref{eq:loss} uses $s{=}20$, $\xi{=}0.1$, $\lambda_{\mathrm{geom}}{=}0.01$ and $\lambda_{\omega}{=}0.001$.
The pair $(s, \xi)$ is the best setting of a sweep over both, across which raw 1-NN accuracy varies by about 1.5\,pp (Supplementary Section~S2).
There is no data augmentation, and image borders are reflect-padded.
An NCA objective is at best on par with the proxy loss (Supplementary Section~S3).

\paragraph{Computational cost}
Section~\ref{sec:exp:cnn} reports the extraction times of our unoptimized PyTorch implementation.
Training one configuration takes 3--20 minutes on a laptop GPU (NVIDIA RTX~2000 Ada), and up to 50 minutes on KTH-TIPS2, whose protocol adds an inner cross-validation.

\subsection{Main ablation results}
\label{sec:exp:results}

Table~\ref{tab:ablation} reports the ablation on the raw descriptor for seed 42.

\begin{table*}[tbp]
\centering
\caption{Controlled component ablation: standardized 1-NN accuracy (\%) on the raw descriptor, with one training seed (42) and one split per benchmark (Section~\ref{sec:exp:significance} repeats Geometry and All joint over seeds). Each learned column learns a different subset of FLEX-LBP parameters, all else being equal. $\mathrm{LBP}^{riu2}$ is the classical grayscale multi-scale $\mathrm{LBP}^{riu2}_{8,1+8,2+16,2}$ (38 dimensions) under the same protocol; it is a same-protocol reference, not a reproduction of published $\chi^2$-matched results, and re-evaluating it with $\chi^2$ or histogram intersection changes each value by at most 3.4\,pp and leaves the TC32 collapse intact. \textbf{Bold}: best; \underline{underlined}: second best. $\Delta_{\mathrm{oracle}}$: best learned configuration minus Fixed; because that configuration is identified on the test split, $\Delta_{\mathrm{oracle}}$ is an upper bound.}
\label{tab:ablation}
\renewcommand{\arraystretch}{1.15}
\setlength{\tabcolsep}{4pt}
\begin{adjustbox}{max width=\linewidth}
\begin{tabular}{@{}lcc cccccc c@{}}
\toprule
& & & \multicolumn{6}{c}{\textbf{Learned configurations}} & \\
\cmidrule(lr){4-9}
\textbf{Dataset} & $\mathbf{LBP}^{\mathbf{riu2}}$ & \textbf{Fixed} & \textbf{Color} & \textbf{Radius} & \textbf{Threshold} & \textbf{Geometry} & \textbf{Mixer+Q} & \textbf{All joint} & $\boldsymbol{\Delta}_{\mathrm{oracle}}$ \\
\midrule
NewBarkTex    & 60.66 & 78.06 & \underline{79.66} & 77.82 & 77.57 & 77.21 & 77.08 & \textbf{81.00} & $+2.94$ \\
KTH-TIPS2     & 62.73 & 72.43 & 71.25 & \textbf{74.37} & 72.77 & \underline{72.93} & 70.24 & 72.77 & $+1.94$ \\
Outex~TC13    & 69.26 & 85.74 & 82.94 & \underline{88.24} & 85.44 & \textbf{88.38} & 79.56 & 88.09 & $+2.64$ \\
Outex~TC31    & 63.60 & 87.35 & 89.49 & 89.78 & \underline{91.40} & \textbf{91.76} & 86.91 & \underline{91.40} & $+4.41$ \\
Parquet       & 37.72 & 46.49 & 48.25 & 47.37 & 59.65 & 55.26 & \underline{64.04} & \textbf{66.67} & $+20.18$ \\
USPtex        & 72.43 & 95.11 & 95.55 & 95.20 & 95.46 & \underline{96.07} & 94.42 & \textbf{96.51} & $+1.40$ \\
\midrule
\multicolumn{10}{@{}l}{\textit{Photometric shift suites (Section~\ref{sec:exp:photometric})}} \\
Outex~TC32    & 13.24 & 58.09 & 61.76 & 57.21 & \textbf{64.85} & 63.75 & \underline{64.56} & 61.47 & $+6.76$ \\
Outex~TC34    & 63.93 & 33.20 & 25.37 & \underline{34.71} & \textbf{35.40} & 32.94 & 26.88 & 26.36 & $+2.20$ \\
\bottomrule
\end{tabular}
\end{adjustbox}
\end{table*}

\paragraph{Component importance depends on the dataset}
Geometry learning (radii, angles and thresholds) is best on TC31 and TC13, and joint optimization is best on NewBarkTex, Parquet and USPtex.
On the photometric suites TC32 and TC34, threshold learning alone is best.
Because TC31, TC32 and TC34 share their training images, their different winners show that the best configuration depends on the test-time shift and not only on the training data.
Radius learning alone improves six of the eight benchmarks, by at most 2.50\,pp (TC13).

\paragraph{Learning color or quantization in isolation is unreliable}
The color transform learned alone improves five of the eight benchmarks but is never the best configuration, and it degrades KTH-TIPS2, TC13 and TC34 (by 7.83\,pp on TC34).
Mixer+Q is even less reliable: it falls below Fixed on six of the eight benchmarks.
Part of this loss may come from its coarser starting quantization, which this ablation does not separate from learning.

\paragraph{A staged schedule does not help}
We also test a sequential schedule that learns the geometry first and the color transform second, and ends with a short joint fine-tuning phase.
It checks whether such a curriculum avoids the failures of joint optimization (Section~\ref{sec:analysis:competing}).
After full postprocessing, it underperforms the best single-pass configuration on every dataset, by 0.6\,pp (NewBarkTex, USPtex) to 9.7\,pp (Parquet), so single-pass training remains the default.

\subsection{Postprocessing pipeline results}
\label{sec:exp:postprocessing}

Supplementary Table~S6 separates the two postprocessing steps behind Table~\ref{tab:endtoend} for seed 42.

\begin{table*}[tbp]
\centering
\caption{End-to-end accuracy (\%) after postprocessing (PP: uniform-u2 projection and power normalization). The descriptors are read out by standardized 1-NN, the protocol of this paper, and by shrinkage LDA. Values are means $\pm$ standard deviations over training seeds: five for Geometry (42--46), three for All joint (42--44), and, for Fixed, the five color initializations of the Geometry seeds. Sig.: number of seeds in which the learned descriptor is significantly better\,/\,worse than the Fixed descriptor of the same seed with the same readout (exact McNemar test, $p{<}0.05$). Oracle: the best of the six learned configurations of Table~\ref{tab:ablation} after postprocessing (seed 42), chosen on the test split and therefore an upper bound. $\mathrm{LBP}^{riu2}$: the classical reference of Table~\ref{tab:ablation}, read out by 1-NN. Mean: over the six main benchmarks; its Sig.\ cells sum the rows above. \textbf{Bold}: best of each readout in each row, the Oracle excepted.}
\label{tab:endtoend}
\renewcommand{\arraystretch}{1.15}
\setlength{\tabcolsep}{3.5pt}
\begin{adjustbox}{max width=\textwidth}
\begin{tabular}{@{}lc cc cccc cccc l@{}}
\toprule
& & \multicolumn{2}{c}{\textbf{Fixed+PP}} & \multicolumn{4}{c}{\textbf{Geometry+PP}} & \multicolumn{4}{c}{\textbf{All joint+PP}} & \\
\cmidrule(lr){3-4}\cmidrule(lr){5-8}\cmidrule(lr){9-12}
\textbf{Dataset} & $\mathbf{LBP}^{\mathbf{riu2}}$ & \textbf{1-NN} & \textbf{LDA} & \textbf{1-NN} & \textbf{Sig.} & \textbf{LDA} & \textbf{Sig.} & \textbf{1-NN} & \textbf{Sig.} & \textbf{LDA} & \textbf{Sig.} & \textbf{Oracle 1-NN} \\
\midrule
NewBarkTex & 60.66 & 83.19 $\pm$ 0.34 & 90.49 $\pm$ 0.45 & 83.31 $\pm$ 0.71 & 0\,/\,0 & 91.54 $\pm$ 0.23 & 0\,/\,0 & \textbf{85.54} $\pm$ 0.88 & 2\,/\,0 & \textbf{93.54} $\pm$ 0.94 & 3\,/\,0 & 84.56 (All joint) \\
KTH-TIPS2 & 62.73 & 75.33 $\pm$ 0.39 & 76.16 $\pm$ 0.34 & \textbf{82.04} $\pm$ 2.70 & 5\,/\,0 & 76.24 $\pm$ 1.99 & 0\,/\,0 & 72.88 $\pm$ 0.21 & 0\,/\,3 & \textbf{76.95} $\pm$ 1.93 & 1\,/\,0 & 84.23 (Geometry) \\
Outex~TC13 & 69.26 & 84.47 $\pm$ 0.64 & 90.94 $\pm$ 0.62 & \textbf{87.53} $\pm$ 0.39 & 5\,/\,0 & 90.29 $\pm$ 0.63 & 0\,/\,1 & 86.57 $\pm$ 0.22 & 2\,/\,0 & \textbf{90.98} $\pm$ 0.47 & 0\,/\,0 & 87.94 (Geometry) \\
Outex~TC31 & 63.60 & 93.38 $\pm$ 0.28 & 93.38 $\pm$ 0.32 & \textbf{96.41} $\pm$ 0.43 & 5\,/\,0 & 94.73 $\pm$ 0.57 & 3\,/\,0 & 95.91 $\pm$ 1.04 & 3\,/\,0 & \textbf{95.95} $\pm$ 1.03 & 3\,/\,0 & 96.40 (Geometry) \\
Parquet & 37.72 & 45.96 $\pm$ 1.00 & 69.30 $\pm$ 2.56 & 52.46 $\pm$ 4.09 & 3\,/\,0 & 72.46 $\pm$ 2.37 & 1\,/\,0 & \textbf{61.40} $\pm$ 1.52 & 3\,/\,0 & \textbf{77.49} $\pm$ 1.34 & 2\,/\,0 & 63.16 (Mixer+Q) \\
USPtex & 72.43 & 94.96 $\pm$ 0.21 & 98.87 $\pm$ 0.09 & 96.82 $\pm$ 0.96 & 4\,/\,0 & 99.39 $\pm$ 0.11 & 2\,/\,0 & \textbf{97.09} $\pm$ 0.79 & 3\,/\,0 & \textbf{99.48} $\pm$ 0.09 & 1\,/\,0 & 97.99 (All joint) \\
\midrule
Mean & 61.07 & 79.55 & 86.52 & 83.10 & 22\,/\,0 & 87.44 & 6\,/\,1 & \textbf{83.23} & 13\,/\,3 & \textbf{89.07} & 10\,/\,0 & 85.71 \\
\midrule
\multicolumn{13}{@{}l}{\textit{Photometric shift suites (Section~\ref{sec:exp:photometric})}} \\
Outex~TC32 & 13.24 & 74.71 $\pm$ 0.65 & \textbf{68.25} $\pm$ 1.11 & \textbf{83.10} $\pm$ 1.10 & 5\,/\,0 & 67.22 $\pm$ 1.35 & 0\,/\,2 & 74.88 $\pm$ 4.62 & 1\,/\,1 & 60.51 $\pm$ 8.81 & 1\,/\,2 & 84.56 (Geometry) \\
Outex~TC34 & \textbf{63.93} & 32.54 $\pm$ 0.75 & \textbf{30.10} $\pm$ 0.19 & 33.83 $\pm$ 1.94 & 2\,/\,1 & 29.75 $\pm$ 0.78 & 0\,/\,1 & 26.74 $\pm$ 3.96 & 0\,/\,3 & 25.49 $\pm$ 2.32 & 0\,/\,3 & 36.10 (Threshold) \\
\bottomrule
\end{tabular}
\end{adjustbox}
\end{table*}

\paragraph{Postprocessing and learning contribute separately}
On the six main benchmarks, postprocessing adds $+2.02$\,pp on average to the Fixed descriptor (seed 42), and learning the geometry adds $+3.55$\,pp on top of it (five seeds).
On TC31 and TC32, uniform-u2 projection alone lifts the Fixed descriptor by $+5.74$ and $+17.20$\,pp (seed 42), more than learning gains on the raw descriptor ($+4.41$\,pp with Geometry and $+6.76$\,pp with Threshold).
The learning gain nevertheless survives postprocessing and is sometimes amplified by it.
On TC32 and KTH-TIPS2, Geometry gains $+5.66$ and $+0.50$\,pp over Fixed before postprocessing (seed 42) and $+8.40$ and $+6.71$\,pp after it (paired over five seeds).
Postprocessing does not always help: u2 projection slightly lowers TC13 for both the Fixed and learned descriptors (seed 42), so the non-uniform codes that it pools sometimes carry discriminative information.

\paragraph{A supervised linear readout helps on natural color textures but not under controlled shifts}
Backbone-based and randomized-network pipelines such as RADAM~\citep{scabini2023radam} and that of \citet{ribas2024randnn} classify with LDA or a linear support vector machine (SVM).
For comparability, the LDA columns of Table~\ref{tab:endtoend} add a shrinkage LDA on the same postprocessed descriptors.
On NewBarkTex, TC13, Parquet and USPtex, LDA is clearly stronger than 1-NN for all three descriptors.
Under LDA, the learning gain shrinks: Geometry is above its paired Fixed descriptor on five of the six main benchmarks, by $+0.92$\,pp on average.
The gain is significant in only 6 of the 30 tests, and Geometry is below Fixed on TC13 ($-0.65$\,pp, significantly in one seed).
All joint keeps more of its gain: over its three seeds, it is above its paired Fixed descriptor on all six main benchmarks, by $+2.68$\,pp on average.
Its gain is significant in 10 of the 18 tests, with no significant loss.
On KTH-TIPS2 and on the suites with a controlled train/test change, LDA brings Fixed and Geometry little or no gain and can cost several points.
With LDA, All joint falls below its paired Fixed descriptor under the photometric shifts (TC32 $-7.87$\,pp, TC34 $-4.63$\,pp).

\subsection{Training seeds and statistical significance}
\label{sec:exp:significance}

Each texture benchmark has a single test split, so we compare two models by the exact McNemar test on their predictions for the same test images.
It accounts for the finite test sample, not for training randomness, which we measure over seeds.
The paragraphs below use the 1-NN columns of Table~\ref{tab:endtoend}; Section~\ref{sec:exp:postprocessing} discusses the LDA columns.

\paragraph{Geometry yields reproducible gains}
Over five seeds, Geometry is more accurate than the paired Fixed descriptor on all six main benchmarks, by $+3.55$\,pp on average (Table~\ref{tab:endtoend}).
The gain is significant in 22 of the 30 seed-level tests, and 17 of these survive a Bonferroni correction ($p{<}0.0017$).
On NewBarkTex, the gain vanishes ($+0.12$\,pp, significant in no seed), and no seed is significantly worse than its Fixed descriptor on any main benchmark.
Single-seed results overstate the gain: after postprocessing, the seed-42 run of Table~\ref{tab:ablation} is the best or second best of the five on seven of the eight benchmarks.
On the illuminant suite TC34, Geometry gains only $+1.29$\,pp, with two seeds significantly better and one significantly worse (Supplementary Section~S13).

\paragraph{All joint can gain more, but not reliably}
All joint gains slightly more on average ($+3.69$\,pp over the six main benchmarks), but less reliably.
It dominates on Parquet ($+15.50$\,pp, significant in all three seeds).
It is, however, significantly worse than the Fixed descriptor in all three seeds on KTH-TIPS2 ($-2.56$\,pp) and TC34 ($-5.68$\,pp).
Of its 18 tests on the main benchmarks, 13 are significant gains and 3 significant losses, against 22 gains and no loss in the 30 tests of Geometry.

\paragraph{Choosing the configuration without the test set}
The oracle results of Tables~\ref{tab:ablation} and~\ref{tab:endtoend} use the configuration that scores best on each test split, so they are optimistic.
Choosing the configuration per dataset by validation 1-NN does not solve this, because the holdouts are too small or too close to the training distribution to rank the configurations.
TC31, TC32 and TC34 share one training partition and therefore have nearly identical validation scores, while their test accuracies differ by more than 50\,pp.
The validation-selected configurations (seed 42) even lose significantly to Fixed on KTH-TIPS2 ($-2.28$\,pp) and TC13 ($-2.94$\,pp).
The robust choice is to apply one configuration, Geometry, unchanged to every benchmark, and we recommend it.
This choice was informed by the seed-42 results of Table~\ref{tab:ablation}, so those results are not an independent test of it.
The four additional seeds and Kather-CRC were run after the choice was made and are such a test.
Over seeds, Geometry keeps its gains, and on Kather-CRC it significantly improves the Fixed descriptor (Section~\ref{sec:exp:kather}).
Color learning is the fragile ingredient: for seed 42, every configuration that learns the color transform loses significantly to Fixed on at least two benchmarks.

\paragraph{Sensitivity to the estimator, the temperature and the templates}
The adjacent-bin estimator of Eq.~\eqref{eq:ste_chain} trains better than the bit-aware alternative of Section~\ref{sec:method:ste} on seven of the eight benchmarks, significantly on three (seed 42).
Among the temperatures $\tau{=}3$, 10 and 30, none is best on all four benchmarks tested.
Swapping the radius templates changes Geometry by $-3.16$ to $+4.39$\,pp but keeps its gain over the same-template Fixed descriptor positive on all eight benchmarks, significantly on five (Supplementary Section~S16).
With the other template, Geometry is significantly less accurate on KTH-TIPS2 ($-3.03$\,pp) and TC32 ($-3.16$\,pp).
Both use the wide template, which was introduced for KTH-TIPS2, so this choice may have benefited from test-set information there.

\subsection{Comparison within the LBP family}
\label{sec:exp:baselines}

Table~\ref{tab:baselines} compares FLEX-LBP with classical and learned members of the LBP family, all rerun under our protocol: the same splits, the same $128{\times}128$ inputs and the same test images.
The classical descriptors are the anchor $\mathrm{LBP}^{riu2}$, the joint CLBP\_S/M/C histogram~\citep{guo2010clbp}, LTP~\citep{tan2010ltp}, MRELBP~\citep{liu2016mrelbp} and BSIF~\citep{kannala2012bsif}, with the default parameters of their papers and nothing tuned on our data.
NLBP~\citep{peeples2024nehd} is trained with the authors' public code and recipe.
PCANet~\citep{chan2015pcanet} follows the texture setting of its paper, with filters learned without labels on each training split (Supplementary Section~S11 details every baseline).
Two fixed variants of our pipeline, both with the identity color transform, complete the table.
Sign-only keeps only the sign codes, which makes it an OCLBP-style descriptor~\citep{maenpaa2004classification} at our three scales, with both orderings of each inter-channel pair, our sampling radii and our postprocessing.
LTP coding replaces the sign and magnitude codes by the upper and lower codes of LTP, with a dead zone of $\pm 0.05$ on intensities in $[0,1]$.

\begin{table*}[tbp]
\centering
\caption{Comparison within the LBP family on the six main benchmarks: accuracy (\%). Our pipeline: postprocessed standardized 1-NN. Classical descriptors: best of four variants (gray or RGB input, $\chi^2$ or standardized 1-NN), chosen on the test split, which favors them; $^{\mathrm{c}}$: the RGB variant is the best one. NLBP: best of its own classifier and 1-NN on its features, local or global aggregation. PCANet: best of twelve variants (gray or RGB input, block or global histograms, $\chi^2$ 1-NN, standardized 1-NN or linear SVM). Sign-only, LTP coding and Fixed: our three-branch, nine-pair color pipeline with fixed parameters and three different codes. Sign-only and LTP coding use the identity color transform, and Fixed is the mean over its five color initializations. Geometry and All joint: means over five and three seeds. $^{\blacktriangle}$/$^{\triangle}$: significantly more accurate than the strongest entry of the column that is not learned FLEX-LBP (classical, PCANet, NLBP or our fixed pipeline) in all/some seeds (exact McNemar test, $p{<}0.05$). \textbf{Bold}: best of each column. NBT: NewBarkTex; KTH: KTH-TIPS2; PAQ: Parquet; USP: USPtex.}
\label{tab:baselines}
\renewcommand{\arraystretch}{1.15}
\setlength{\tabcolsep}{4.5pt}
\begin{adjustbox}{max width=\textwidth}
\begin{tabular}{@{}l cccccc c@{}}
\toprule
\textbf{Method} & \textbf{NBT} & \textbf{KTH} & \textbf{TC13} & \textbf{TC31} & \textbf{PAQ} & \textbf{USP} & \textbf{Mean} \\
\midrule
$\mathrm{LBP}^{riu2}$ & 60.66 & 65.18 & 72.65 & 63.60 & 40.35 & 72.86 & 62.55 \\
CLBP & 76.96$^{\mathrm{c}}$ & 69.56$^{\mathrm{c}}$ & 82.21$^{\mathrm{c}}$ & 86.32$^{\mathrm{c}}$ & 62.28$^{\mathrm{c}}$ & 93.72$^{\mathrm{c}}$ & 78.51 \\
LTP & 71.81$^{\mathrm{c}}$ & 68.13$^{\mathrm{c}}$ & 72.35$^{\mathrm{c}}$ & 82.13$^{\mathrm{c}}$ & 56.14$^{\mathrm{c}}$ & 91.54$^{\mathrm{c}}$ & 73.68 \\
MRELBP & 69.61$^{\mathrm{c}}$ & 74.54$^{\mathrm{c}}$ & 77.35$^{\mathrm{c}}$ & 75.96$^{\mathrm{c}}$ & 46.49$^{\mathrm{c}}$ & 90.14$^{\mathrm{c}}$ & 72.35 \\
BSIF & 80.39$^{\mathrm{c}}$ & 65.60$^{\mathrm{c}}$ & 81.91$^{\mathrm{c}}$ & 90.07$^{\mathrm{c}}$ & 32.46$^{\mathrm{c}}$ & 90.14$^{\mathrm{c}}$ & 73.43 \\
\midrule
NLBP (learned) & 66.79 & 61.05 & 55.88 & 91.62 & 33.33 & 67.89 & 62.76 \\
PCANet (learned) & 80.64$^{\mathrm{c}}$ & 70.74$^{\mathrm{c}}$ & 82.50$^{\mathrm{c}}$ & 89.93$^{\mathrm{c}}$ & 50.88$^{\mathrm{c}}$ & 96.07$^{\mathrm{c}}$ & 78.46 \\
\midrule
Sign-only (OCLBP-style) & 77.57 & 71.50 & 85.44 & 93.68 & 42.11 & 92.76 & 77.18 \\
LTP coding & 78.31 & 74.87 & \textbf{88.68} & \textbf{97.28} & \textbf{64.04} & 96.16 & 83.22 \\
Fixed (SM-LBP) & 83.19 & 75.33 & 84.47 & 93.38 & 45.96 & 94.96 & 79.55 \\
\midrule
FLEX-LBP Geometry & 83.31 & \textbf{82.04$^{\blacktriangle}$} & 87.53 & 96.41 & 52.46 & 96.82$^{\triangle}$ & 83.10 \\
FLEX-LBP All joint & \textbf{85.54$^{\triangle}$} & 72.88 & 86.57 & 95.91 & 61.40 & \textbf{97.09$^{\triangle}$} & \textbf{83.23} \\
\bottomrule
\end{tabular}
\end{adjustbox}
\end{table*}

Against the external descriptors, FLEX-LBP is the most accurate on five of the six main benchmarks (Table~\ref{tab:baselines}).
Geometry is significantly more accurate than the strongest classical descriptor in all five seeds on KTH-TIPS2 (MRELBP), TC13 (CLBP) and TC31 (BSIF), and in four seeds on USPtex (CLBP).
On NewBarkTex, it is 2.9\,pp ahead of BSIF, but significantly so in one seed only.
On Parquet, CLBP on RGB is ahead of Geometry, significantly in two seeds, and level with All joint.
Among the learned descriptors, PCANet is ahead of NLBP on average but below Geometry on all six benchmarks.
Geometry leads it significantly in all five seeds on KTH-TIPS2, TC13 and TC31, in two on USPtex and in one on NewBarkTex, with a tie on Parquet.
NLBP stays far below on every benchmark, and Geometry beats it in all five seeds on all six benchmarks.
Averaged over the six benchmarks, Geometry reaches 83.10\%, against 78.51\% for the best classical descriptor (CLBP), 78.46\% for PCANet and 62.76\% for NLBP.

To check whether the ranking depends on the 1-NN readout of the classical descriptors, we apply the shrinkage LDA of Section~\ref{sec:method:postprocessing} to every descriptor (Supplementary Section~S22).
Within the LBP family, FLEX-LBP is again the most accurate on five of the six benchmarks, but not the same five.
On NewBarkTex, TC13, TC31, Parquet and USPtex, even Fixed stays above the strongest classical descriptor and PCANet.
Geometry is significantly ahead of the strongest classical descriptor in all five seeds on four of these benchmarks and in two seeds on Parquet.
All joint is significantly ahead in all three seeds on all five.
KTH-TIPS2 is the exception: MRELBP reaches 86.76\%, significantly above Geometry (76.24\%) and All joint (76.95\%) in every seed, which reverses the 1-NN result of Geometry.

Within our own pipeline, the coding matters as much as learning.
With fixed LTP codes in place of the sign and magnitude codes, the pipeline reaches the same mean accuracy as Geometry (83.22\% against 83.10\%).
LTP coding is ahead on TC13, TC31 and Parquet, significantly in two and three of the five seeds on TC31 and Parquet.
It is behind on NewBarkTex, KTH-TIPS2 and USPtex, significantly in all five seeds on the first two and in three on USPtex.

The gap between LTP coding and our sign-plus-magnitude coding lies mainly in the type of magnitude threshold (Supplementary Section~S17).
Our magnitude code compares each magnitude with the mean magnitude of the same image (Section~\ref{sec:method:smlbp}), which discards the absolute size of the differences that the absolute dead zone of LTP keeps.
On TC31 and Parquet, this size is itself a strong class cue.
Used alone, the 27 per-image mean magnitudes $\mu_{b}^{(i,j)}$ of the Fixed descriptor reach 87.57\% on TC31 and 74.56\% on Parquet, more on Parquet than any descriptor of Table~\ref{tab:baselines}.
With an absolute magnitude threshold equal to the dead zone of LTP coding ($m{>}0.05$), the Fixed descriptor with the identity color transform rises from 79.42\% to 83.20\% on average.
It thus reaches the level of LTP coding and closes most of the gap to it on TC13, TC31 and Parquet.
Conversely, with a dead zone set to the mean magnitude of each image, LTP coding loses most of its lead (Parquet 47.37\% instead of 64.04\%).
Restricting the descriptor to one kind of channel pair shows that on Parquet and TC13, the gain of the absolute threshold comes from the inter-channel pairs (Supplementary Table~S18).
Their magnitudes include the difference between two channels, that is, color.
For the whole descriptor, the absolute threshold loses accuracy on NewBarkTex and under the illuminant change of TC34, so neither type of threshold is best on every benchmark.

\subsection{Photometric shifts: noise and illuminant}
\label{sec:exp:photometric}

On the two photometric suites, the color descriptor is far behind the best descriptors, PCANet and BSIF (Table~\ref{tab:photometric}).
To locate the failure, we pair every test image with its training twin, the same crop acquired under the training conditions.
We then dissect the Fixed descriptor by branch and by channel pair (Supplementary Section~S12).

\begin{table}[tbp]
\centering
\small
\caption{Photometric shift suites: postprocessed standardized 1-NN accuracy (\%). Classical descriptors, PCANet and NLBP as in Table~\ref{tab:baselines} (best variant on the test split; $^{\mathrm{c}}$: RGB). Fixed with RGB input: the single run to which the remedies below are applied. Geometry with RGB input: mean $\pm$ SD over five seeds. The remedies are applied at inference, without retraining, to the Fixed descriptor and to the trained Geometry model (seed 42). $\Delta_{\mathrm{main}}$: the mean accuracy change that the remedy causes for the Fixed descriptor on the six main benchmarks (Supplementary Section~S12). The remedies were chosen by diagnosing these two suites, so they are not held-out results. \textbf{Bold}: best of each accuracy column.}
\label{tab:photometric}
\renewcommand{\arraystretch}{1.12}
\setlength{\tabcolsep}{4pt}
\begin{adjustbox}{max width=\linewidth}
\begin{tabular}{@{}lccc@{}}
\toprule
\textbf{Descriptor} & \textbf{TC32 (noise)} & \textbf{TC34 (illuminant)} & $\boldsymbol{\Delta}_{\mathrm{main}}$ \\
\midrule
$\mathrm{LBP}^{riu2}$ & 13.24 & 63.93 & -- \\
CLBP & 24.63 & 73.71 & -- \\
LTP & 16.03 & 55.81 & -- \\
MRELBP & 90.15 & 87.57 & -- \\
BSIF & 97.72 & 97.61 & -- \\
NLBP (learned) & 44.12 & 28.79 & -- \\
PCANet (learned) & \textbf{98.53}$^{\mathrm{c}}$ & \textbf{98.46}$^{\mathrm{c}}$ & -- \\
\midrule
FLEX-LBP Fixed, RGB input & 74.12 & 33.49 & -- \\
FLEX-LBP Geometry, RGB input & 83.10 $\pm$ 1.10 & 33.83 $\pm$ 1.94 & -- \\
\midrule
\multicolumn{4}{@{}l}{\textit{Fixed (seed 42) with an input-side remedy}} \\
\quad median support $3{\times}3$ & 96.10 & 33.12 & $-1.55$ \\
\quad median-filtered input & 95.07 & 33.16 & $-1.22$ \\
\quad gray-world input & 55.15 & 96.99 & $+0.15$ \\
\quad gray-world + median support & 92.35 & 96.07 & $-0.76$ \\
\quad luminance input & 66.03 & 93.38 & $-12.22$ \\
\multicolumn{4}{@{}l}{\textit{Geometry (seed 42) with the same remedy, without retraining}} \\
\quad median support $3{\times}3$ & 97.43 & -- & -- \\
\quad median-filtered input & 98.09 & -- & -- \\
\quad gray-world input & -- & 92.21 & -- \\
\quad gray-world + median support & 89.12 & 92.90 & -- \\
\quad luminance input & -- & 80.59 & -- \\
\bottomrule
\end{tabular}
\end{adjustbox}
\end{table}

\paragraph{Noise: single-pixel comparisons drown}
In TC32, the added noise has an SD of 5.1 gray levels per channel.
This is larger than the typical contrast between neighboring pixels of the clean images (a mean absolute difference of 4.3 gray levels), so many single-pixel comparisons become random.
The fine branch suffers most, and the intra-channel pairs (54.56\%) more than the inter-channel pairs (75.29\%), whose differences between channels are often larger than the noise.
Learned thresholds help, as an offset that ignores small differences should, but averaging before comparing helps much more.
Sampling each neighbor and the center as the median of a $3{\times}3$ patch, the regional sampling of MRELBP, raises the Fixed descriptor to 96.10\%.
A $3{\times}3$ median-filtered input raises it to 95.07\%.

\paragraph{Illuminant: comparisons between channels flip}
In TC34, an affine fit between each test image and its training twin shows that the illuminant scales the three channels by different gains (Supplementary Section~S12).
Comparisons within one channel are invariant to such a gain, but comparisons between channels are not.
On its intra-channel pairs alone, the Fixed descriptor reaches 79.12\%, on its inter-channel pairs alone 21.36\% and on a luminance input 93.38\%.
Learned thresholds also break the gain invariance of the intra-channel sign codes (40.66\% with Geometry against 77.79\% for Fixed, seed 42).
Normalizing each channel by its image mean, the gray-world assumption~\citep{buchsbaum1980grayworld}, cancels per-channel gains and raises the Fixed descriptor to 96.99\%.
This agrees with \citet{maenpaa2004classification}: under the same test illuminants (Outex~TC14), opponent color texture descriptors were hit especially hard, and a comprehensive color normalization helped considerably.

\paragraph{No remedy is free}
Each remedy fixes its own nuisance and can hurt elsewhere (Table~\ref{tab:photometric}; Supplementary Section~S12 reports its cost on every benchmark).
On the six main benchmarks, the median support costs most on Parquet ($-7.01$\,pp), where fine wood grain separates the classes.
Luminance input costs 12.2\,pp on average, which measures what color contributes on these benchmarks.
Gray-world input combined with the median support handles both suites at an average cost of 0.76\,pp on the main benchmarks.
With its remedy, the Fixed descriptor is statistically tied with BSIF on TC34 (96.99\% against 97.61\%, $p{=}0.12$) and slightly behind it on TC32 (96.10\% against 97.72\%, $p{=}0.02$).
PCANet stays significantly ahead of the remedied Fixed descriptor on both suites ($p{<}10^{-3}$).

Plugged into the trained Geometry model at inference, without retraining, the median remedies keep the learning gain on TC32 (Table~\ref{tab:photometric}).
With the median support and the median-filtered input, Geometry is significantly above the Fixed descriptor with the same remedy ($p{=}0.005$ and $p{<}10^{-9}$) and statistically tied with BSIF ($p{=}0.71$ and $0.59$).
With the median-filtered input, it is also statistically tied with PCANet ($p{=}0.42$).
The per-channel normalization does not keep it: on TC34, Geometry with gray-world input reaches 92.21\%, below the Fixed descriptor with the same input (96.99\%), because its thresholds were learned on unnormalized intensities.

\subsection{Application to colorectal cancer histology}
\label{sec:exp:kather}

On Kather-CRC, we run Fixed, Geometry and All joint with the USPtex configuration and one training run per fold, and the other descriptors exactly as in Section~\ref{sec:exp:baselines} (Supplementary Section~S15).
Table~\ref{tab:kather} reports the results over the five folds, and the McNemar tests below pool their predictions (5,000 test tiles).

\begin{table}[tbp]
\centering
\small
\caption{Kather-CRC colorectal cancer histology (eight tissue classes, stratified five-fold cross-validation): accuracy (\%), mean $\pm$ standard deviation over folds. Classical descriptors, PCANet and NLBP as in Table~\ref{tab:baselines} (classical: best of $\chi^2$ and standardized 1-NN; $^{\mathrm{c}}$: RGB input). FLEX-LBP descriptors after postprocessing, one training run per fold. LDA: the shrinkage LDA of Section~\ref{sec:method:postprocessing} on every descriptor, with the same input as its 1-NN variant; PCANet: its best variant in each column (1-NN column: linear SVM on RGB block histograms; LDA: global RGB histograms, $2{,}048$ dimensions); NLBP: LDA on its 48 pooled histogram features. \textbf{Bold}: best of each accuracy column. For reference, \citet{kather2016multi} reported 87.4\% with combined gray-level descriptors and an SVM under ten-fold cross-validation.}
\label{tab:kather}
\renewcommand{\arraystretch}{1.12}
\setlength{\tabcolsep}{4pt}
\begin{adjustbox}{max width=\linewidth}
\begin{tabular}{@{}lccr@{}}
\toprule
\textbf{Descriptor} & \textbf{1-NN} & \textbf{LDA} & \textbf{Dims} \\
\midrule
$\mathrm{LBP}^{riu2}$ & 73.78 $\pm$ 1.38 & 72.84 $\pm$ 1.00 & 38 \\
CLBP, gray & 89.54 $\pm$ 1.28 & 85.48 $\pm$ 1.12 & 2{,}200 \\
CLBP, RGB & 92.80 $\pm$ 0.80 & 88.28 $\pm$ 1.13 & 6{,}600 \\
LTP$^{\mathrm{c}}$ & 89.92 $\pm$ 1.02 & 89.02 $\pm$ 1.42 & 324 \\
MRELBP$^{\mathrm{c}}$ & 89.28 $\pm$ 1.75 & 90.44 $\pm$ 1.66 & 2{,}400 \\
BSIF$^{\mathrm{c}}$ & 84.36 $\pm$ 1.70 & 87.42 $\pm$ 1.12 & 2{,}304 \\
NLBP (learned) & 84.04 $\pm$ 1.74 & 76.98 $\pm$ 2.01 & 48 \\
PCANet (learned)$^{\mathrm{c}}$ & 85.22 $\pm$ 0.73 & 87.06 $\pm$ 1.21 & 32{,}768 \\
\midrule
FLEX-LBP Fixed & 92.18 $\pm$ 0.98 & 93.98 $\pm$ 1.03 & 3{,}186 \\
FLEX-LBP Geometry & 93.28 $\pm$ 0.63 & 93.86 $\pm$ 1.02 & 3{,}186 \\
FLEX-LBP All joint & \textbf{93.82} $\pm$ 0.83 & \textbf{94.50} $\pm$ 1.04 & 3{,}186 \\
\bottomrule
\end{tabular}
\end{adjustbox}
\end{table}

Learning improves the Fixed descriptor under the standardized 1-NN readout.
Geometry gains $+1.10$\,pp ($p{=}3.9{\times}10^{-5}$) and All joint $+1.64$\,pp ($p{=}1.9{\times}10^{-9}$), and All joint is also significantly more accurate than Geometry ($p{=}0.04$).
All joint reaches $93.82 \pm 0.83\%$, significantly above the strongest classical descriptor, CLBP on the RGB channels with $\chi^2$ matching ($92.80 \pm 0.80\%$, $p{=}0.014$).
Geometry ($p{=}0.27$) and the Fixed descriptor ($p{=}0.16$) are statistically tied with CLBP, and the learned PCANet and NLBP stay far below ($p{<}10^{-50}$ against both learned FLEX-LBP configurations).
With LDA, the three FLEX-LBP descriptors reach 93.9--94.5\%, and the learning gain is no longer significant (Geometry $-0.12$\,pp, $p{=}0.70$; All joint $+0.52$\,pp, $p{=}0.07$).
The other descriptors stay below them with this readout, which lowers CLBP on the RGB channels and NLBP (Table~\ref{tab:kather}).
The three FLEX-LBP descriptors are significantly more accurate than the strongest classical descriptor with LDA, MRELBP on the RGB channels (90.44\%), and than PCANet and NLBP ($p{<}10^{-17}$).

Color matters on this benchmark: CLBP is 3.3\,pp more accurate on the RGB channels than on the gray image (Table~\ref{tab:kather}).
Gray-world input, which rescued TC34, costs the Fixed descriptor 3.0\,pp here (Supplementary Section~S15).
Stain color is informative and constant within this single-center benchmark, so normalizing it away removes information.

\subsection{Comparison with frozen pretrained CNN features}
\label{sec:exp:cnn}

Frozen features of ImageNet-pretrained CNNs are a standard deep reference in color texture classification~\citep{cimpoi2016deep,bellocerezo2019comparative}.
In the comparison of \citet{bellocerezo2019comparative}, the globally pooled ResNet-50 features that we use ranked first among 68 hand-crafted and CNN-based descriptors.
We extract them, without fine-tuning, from a ResNet-50~\citep{he2016resnet} with its original ImageNet weights, trained on the 1.28 million labeled images of ImageNet~\citep{russakovsky2015imagenet}.
We resize each $128{\times}128$ input image to $224{\times}224$ and read out the $2{,}048$-dimensional pooled features by the same standardized 1-NN and shrinkage LDA as FLEX-LBP, on the same splits and test images.
Supplementary Section~S18 details the extraction and shows that newer ImageNet weights and the native $128{\times}128$ input are less accurate on average with both readouts.
Tables~\ref{tab:cnn} and~\ref{tab:cnncost} report the accuracy and the cost of both descriptors.

\begin{table*}[tbp]
\centering
\caption{Frozen ImageNet ResNet-50 features and FLEX-LBP under the same protocol: accuracy (\%) on the same images, splits and test images, with the same readouts. ResNet-50: a single run, since the frozen network has no training seed. Fixed: mean over its five color initializations; Geometry and All joint: means over five and three training seeds. Kather-CRC: mean over the five folds; Mean: over the six main benchmarks. $^{\ast}$:~with the input-side remedy of Section~\ref{sec:exp:photometric} applied at inference without retraining (a $3{\times}3$ median-filtered input on TC32 and a gray-world input on TC34). For FLEX-LBP, these are the seed-42 models of Table~\ref{tab:photometric}; seed 42 is also the most accurate Geometry seed on TC32 without a remedy (84.56\% against a mean of 83.10\%). The remedies were chosen after diagnosing these two suites, so they are not held-out results. \textbf{Bold}: best of each column within each block. ``--'': not reported. NBT: NewBarkTex; KTH: KTH-TIPS2; PAQ: Parquet; USP: USPtex; Kather: Kather-CRC.}
\label{tab:cnn}
\renewcommand{\arraystretch}{1.12}
\setlength{\tabcolsep}{3.5pt}
\begin{adjustbox}{max width=\textwidth}
\begin{tabular}{@{}l ccccccc cc c@{}}
\toprule
& \multicolumn{7}{c}{\textbf{Main benchmarks}} & \multicolumn{2}{c}{\textbf{Photometric}} & \\
\cmidrule(lr){2-8}\cmidrule(lr){9-10}
\textbf{Descriptor} & \textbf{NBT} & \textbf{KTH} & \textbf{TC13} & \textbf{TC31} & \textbf{PAQ} & \textbf{USP} & \textbf{Mean} & \textbf{TC32} & \textbf{TC34} & \textbf{Kather} \\
\midrule
\multicolumn{11}{@{}l}{\textit{Standardized 1-NN}} \\
\quad ResNet-50 (ImageNet, frozen) & 81.00 & \textbf{87.52} & 85.44 & 88.82 & 43.86 & \textbf{98.43} & 80.85 & \textbf{92.57} & \textbf{72.61} & 88.50 \\
\quad FLEX-LBP Fixed & 83.19 & 75.33 & 84.47 & 93.38 & 45.96 & 94.96 & 79.55 & 74.71 & 32.54 & 92.18 \\
\quad FLEX-LBP Geometry & 83.31 & 82.04 & \textbf{87.53} & \textbf{96.41} & 52.46 & 96.82 & 83.10 & 83.10 & 33.83 & 93.28 \\
\quad FLEX-LBP All joint & \textbf{85.54} & 72.88 & 86.57 & 95.91 & \textbf{61.40} & 97.09 & \textbf{83.23} & 74.88 & 26.74 & \textbf{93.82} \\
\midrule
\multicolumn{11}{@{}l}{\textit{Shrinkage LDA}} \\
\quad ResNet-50 (ImageNet, frozen) & \textbf{94.00} & \textbf{93.93} & 90.15 & 90.00 & 52.63 & \textbf{99.74} & 86.74 & \textbf{89.49} & \textbf{70.22} & 93.30 \\
\quad FLEX-LBP Fixed & 90.49 & 76.16 & 90.94 & 93.38 & 69.30 & 98.87 & 86.52 & 68.25 & 30.10 & 93.98 \\
\quad FLEX-LBP Geometry & 91.54 & 76.24 & 90.29 & 94.73 & 72.46 & 99.39 & 87.44 & 67.22 & 29.75 & 93.86 \\
\quad FLEX-LBP All joint & 93.54 & 76.95 & \textbf{90.98} & \textbf{95.95} & \textbf{77.49} & 99.48 & \textbf{89.07} & 60.51 & 25.49 & \textbf{94.50} \\
\midrule
\multicolumn{11}{@{}l}{\textit{With the input-side remedy of Section~\ref{sec:exp:photometric}, standardized 1-NN}} \\
\quad ResNet-50 (ImageNet, frozen) & -- & -- & -- & -- & -- & -- & -- & 93.97$^{\ast}$ & 82.43$^{\ast}$ & -- \\
\quad FLEX-LBP Fixed (seed 42) & -- & -- & -- & -- & -- & -- & -- & 95.07$^{\ast}$ & \textbf{96.99}$^{\ast}$ & -- \\
\quad FLEX-LBP Geometry (seed 42) & -- & -- & -- & -- & -- & -- & -- & \textbf{98.09}$^{\ast}$ & 92.21$^{\ast}$ & -- \\
\bottomrule
\end{tabular}
\end{adjustbox}
\end{table*}

\begin{table*}[tbp]
\centering
\caption{Cost of the descriptors of Table~\ref{tab:cnn}. Learned parameters: parameters of the deployed descriptor fitted to labeled images. The ImageNet classifier of ResNet-50 and the class proxies of the FLEX-LBP training loss are not counted. In every case, the readout is also fitted on the training images of each benchmark (Table~\ref{tab:datasets}). FLEX-LBP stores 90 parameters (Section~\ref{sec:method}), of which Geometry learns 60 and All joint 87 (all but the number of quantization levels of each branch). Because the eight angle parameters of a branch carry seven degrees of freedom (Section~\ref{sec:method:geometry}), these amount to 57 and 84 degrees of freedom. Training images: the labeled images on which these parameters are fitted (``--'': none). Input: input image size in pixels; Dims: descriptor dimension; Size: the stored parameters in single precision. Times per input image on one laptop: one CPU thread with single images, and its GPU (NVIDIA RTX~2000 Ada) in batches of 32 (FLEX-LBP) or 64 (ResNet-50). Supplementary Section~S18 details both measurements.}
\label{tab:cnncost}
\renewcommand{\arraystretch}{1.12}
\setlength{\tabcolsep}{4pt}
\begin{adjustbox}{max width=\textwidth}
\begin{tabular}{@{}l r r c r r r r@{}}
\toprule
\textbf{Descriptor} & \textbf{Learned param.} & \textbf{Training images} & \textbf{Input} & \textbf{Dims} & \textbf{Size} & \textbf{CPU (ms)} & \textbf{GPU (ms)} \\
\midrule
ResNet-50 (ImageNet, frozen) & $23{,}508{,}032$ & $1{,}281{,}167$ (ImageNet) & $224{\times}224$ & $2{,}048$ & 94\,MB & 124 & 3.1 \\
FLEX-LBP Fixed & 0 & -- & $128{\times}128$ & $3{,}186$ & 360\,B & 35--40 & 3 \\
FLEX-LBP Geometry & 60 & 114--$4{,}000$ & $128{\times}128$ & $3{,}186$ & 360\,B & 35--40 & 3 \\
FLEX-LBP All joint & 87 & 114--$4{,}000$ & $128{\times}128$ & $3{,}186$ & 360\,B & 35--40 & 3 \\
\bottomrule
\end{tabular}
\end{adjustbox}
\end{table*}

\paragraph{Neither descriptor is uniformly better}
Averaged over the six main benchmarks, ResNet-50 reaches 80.85\% with 1-NN and 86.74\% with LDA, against 83.10\% and 87.44\% for Geometry and 83.23\% and 89.07\% for All joint.
With LDA, even the untrained Fixed descriptor is at its level (86.52\%).
The LDA difference, however, comes from Parquet: on the other five benchmarks, ResNet-50 is ahead with LDA (93.56\% against 90.44\% for Geometry and 91.38\% for All joint).
With 1-NN, Geometry stays ahead on these five (89.22\% against 88.24\%), mainly because of TC31.
Choosing the best of six ResNet-50 1-NN variants on each test split, as for the classical descriptors, yields a mean of 82.50\%, still below Geometry and All joint (Supplementary Section~S18).

Significant differences occur in both directions (Supplementary Section~S18 counts them per seed).
With 1-NN, ResNet-50 is significantly more accurate than Geometry on KTH-TIPS2 in four of the five seeds and on USPtex in two.
It is statistically tied with Geometry in every seed on NewBarkTex and TC13.
Conversely, it is significantly less accurate on TC31 in all five seeds and on Parquet in one.
All joint is significantly ahead of it in all three seeds on Parquet and NewBarkTex.
With LDA on the main benchmarks, ResNet-50 is significantly more accurate than Geometry on KTH-TIPS2 in all five seeds and on NewBarkTex in three.
It is also significantly more accurate than All joint on KTH-TIPS2 in all three seeds.
Both FLEX-LBP descriptors are significantly ahead on TC31 and Parquet in every seed, and the other tests are ties.
The two largest of these differences thus point in opposite directions: the CNN features lead on KTH-TIPS2, whose test images come from an unseen physical sample, and FLEX-LBP leads on Parquet.
This pattern agrees with the large comparison of \citet{bellocerezo2019comparative}.
There, off-the-shelf CNN features were markedly better on average, especially on non-stationary textures and under several acquisition changes, whereas hand-crafted descriptors were better on stationary textures under steady imaging conditions.

\paragraph{Photometric shifts}
Without a remedy, the CNN features are far more robust to the two photometric shifts than our color descriptor, and significantly more accurate than Geometry in every seed on both suites.
With the same input remedies, the order reverses (Table~\ref{tab:cnn}, rows marked $^{\ast}$).
With a median-filtered input, ResNet-50 is significantly less accurate on TC32 than the Geometry model of seed 42 with the same input ($p{=}4.5{\times}10^{-8}$) and statistically tied with the Fixed descriptor ($p{=}0.18$).
With a gray-world input, it is far less accurate on TC34 than the Fixed descriptor and the Geometry model with the same input ($p{<}10^{-34}$).
Its best input on this suite, luminance, also stays below them (85.99\%, Supplementary Section~S18).
The CNN features thus absorb part of each nuisance without a remedy but gain much less from it.
The difference arises because the failures of our descriptor lie in specific comparisons that the remedies repair (Section~\ref{sec:exp:photometric}).

\paragraph{Histology}
On Kather-CRC, the three FLEX-LBP descriptors are significantly more accurate than ResNet-50 with 1-NN ($p{<}10^{-12}$).
With LDA, All joint is significantly more accurate ($p{=}6.6{\times}10^{-4}$), whereas Geometry and the Fixed descriptor are statistically tied with ResNet-50 ($p{=}0.13$ and $0.06$).

\paragraph{Cost}
The two descriptors differ by orders of magnitude in parameters, training data and storage, but much less in extraction time (Table~\ref{tab:cnncost}).
The CNN needs about 4 billion multiply--adds per $224{\times}224$ image~\citep{he2016resnet}.
At the native $128{\times}128$ input, which is 0.8--0.9\,pp less accurate on average (Supplementary Section~S18), it would need about a third of them.

A frozen ImageNet ResNet-50 is thus a strong reference on natural material images and, without input remedies, under photometric shifts.
It does not, however, dominate an LBP whose 60 to 87 parameters are learned from 114 to $4{,}000$ images.
Fine-tuned and larger backbones are outside this comparison (Section~\ref{sec:analysis:limitations}).

\section{Analysis and discussion}
\label{sec:analysis}

\subsection{Which components matter where, and why}
\label{sec:analysis:components}

The training dynamics partly explain the dataset-dependent results of Section~\ref{sec:exp:results}.
On the NewBarkTex All joint run, the mean gradient norms of the color transform and the thresholds are about 290 and 275 times that of the radii (Supplementary Table~S3).
These norms, not corrected for tensor size or units, measure sensitivity rather than step size, since Adam normalizes the step of each parameter.
Both components act directly on the intensities that feed all $K{\times}9 = 72$ channel-pair comparisons of a pixel.
Small changes to them therefore reshape the histograms substantially, consistent with the disruptive effect of isolated color learning.
The radius acts only through the bilinear interpolation of the sampling positions, where sub-pixel shifts change the sampled intensities little.
Radius learning therefore appears limited less by the importance of scale than by the weakness of the signal that reaches it.

\subsection{Learned parameters are interpretable and reproducible}
\label{sec:analysis:geometry}

\begin{figure*}[tbp]
\centering
\includegraphics[width=0.96\textwidth]{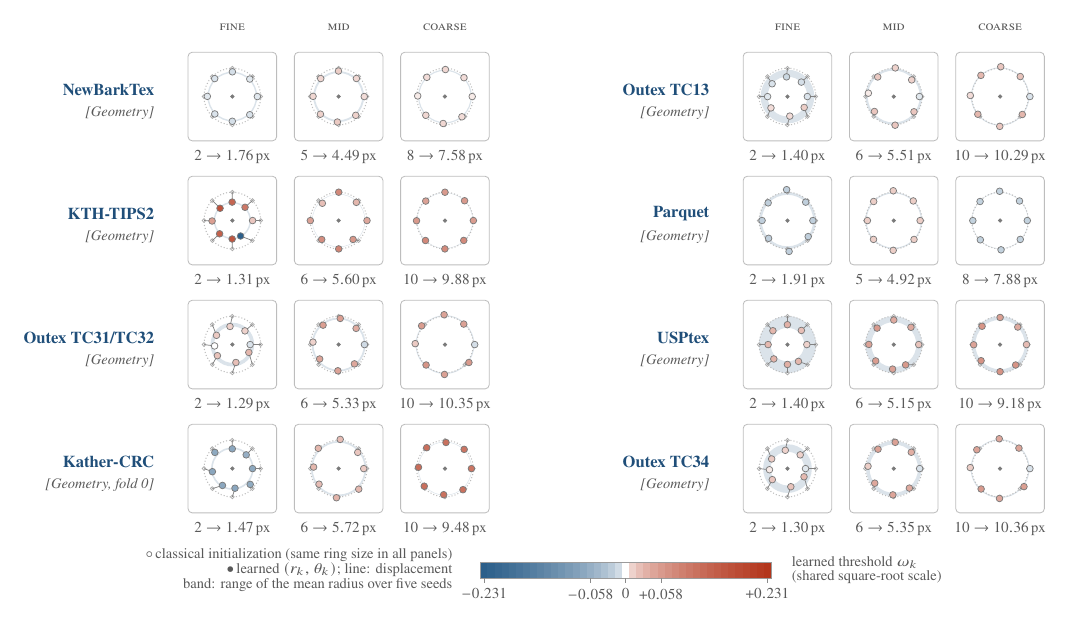}
\caption{Learned LBP sampling geometry of the Geometry configuration (seed 42) on every benchmark. TC31 and TC32 share their training images and schedule, so they share one model; the bottom-left panel shows Kather-CRC (fold 0). Dotted ring and open circles: classical 8-neighbor initialization, scaled to the same size in every panel; filled dots: learned positions at radii $r_{b,k}$ and angles $\theta_{b,k}$, joined to their initial positions by thin lines and colored by the signed threshold $\omega_{b,k}$ on a square-root scale shared by all panels (red positive, blue negative); shaded band: range of the learned mean radius of the branch over the five seeds (Kather-CRC: over the five folds). The labels show the initial and learned mean radius of each branch in pixels for the plotted run.}
\label{fig:geometry}
\end{figure*}

\begin{figure}[tbp]
\centering
\includegraphics[width=\codefigwidth]{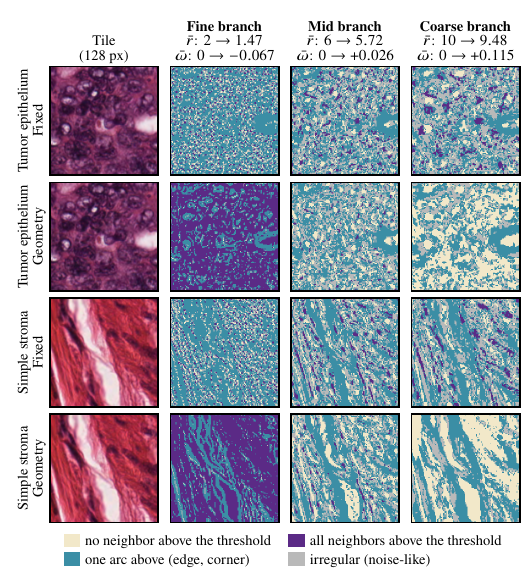}
\caption{Sign codes of the green--green channel pair on two Kather-CRC tiles (the first two of Fig.~\ref{fig:samples}b), computed by the Fixed and the Geometry model of fold~0 with their own radii, angles and thresholds (column heads: mean radius $\bar r$ in pixels and mean threshold $\bar\omega$ of each branch, Fixed $\to$ Geometry). Each 8-bit code is grouped by its circular structure: no neighbor, one contiguous arc of neighbors (edge or corner) or all neighbors above the threshold; or an irregular code with more than two transitions. With $\omega{=}0$, these groups are bright spots or flat regions; edges and corners; and dark spots. A negative learned threshold (fine branch) turns flat regions into `all above', and a positive one (coarse branch) turns them into `none above'.}
\label{fig:codes}
\end{figure}

\paragraph{The direction of adaptation recurs across seeds}
Fig.~\ref{fig:geometry} decodes the sampling geometry that Geometry learns on every benchmark (seed 42).
The coarse radius grows in all five seeds on the Outex training set shared by TC31, TC32 and TC34 (10.00 to $10.21 \pm 0.16$ pixels with the 12-epoch schedule; Supplementary Section~S14).
It shrinks in all five seeds on NewBarkTex, KTH-TIPS2 and Parquet, in all five folds of Kather-CRC (10.00 to $9.39 \pm 0.08$) and in four of five seeds on USPtex.
TC13 has no consistent direction (three of five seeds grow).
The fine radius shrinks in every run but one, which may reflect a preference of the training objective as much as the data (Supplementary Section~S4).
A coarse radius that grows on one training set and shrinks on others cannot be explained that way.
The mean threshold grows from the fine to the coarse branch on every benchmark except KTH-TIPS2, whose fine branch learns the largest thresholds.
On Kather-CRC tiles, the learned thresholds remove most irregular, noise-like codes and specialize the branches (Fig.~\ref{fig:codes}; Supplementary Section~S7).

\paragraph{The adaptation stays mild}
Angular spacing departs from uniform by at most about $19^\circ$ on average over seeds, and the longest sampling radius of a branch exceeds its shortest by about 20\% at most.
This closeness to the initialization is partly by design (bounded radii, $\mathcal{L}_{\mathrm{geom}}$), so the informative finding is the direction of adaptation rather than its size.
Because all these quantities are explicit, bounded parameters, a practitioner can inspect whether each learned change is physically plausible, which end-to-end learned alternatives do not allow.

\paragraph{Color}
The learned color transforms tell a parallel story (Supplementary Fig.~S1).
The coarse branch of the USPtex Mixer+Q run becomes opponent-color-like, and the fine branch of All joint acts as a crude stain axis in every Kather-CRC fold.

\subsection{Limits of learning and of the 1-NN readout}
\label{sec:analysis:competing}

\paragraph{When joint optimization hurts}
Learning more parameters can hurt: All joint is below Threshold on TC32 and TC34 (Table~\ref{tab:ablation}) and, with 1-NN, significantly below the Fixed descriptor on KTH-TIPS2 and TC34 in all three seeds (Table~\ref{tab:endtoend}).
Their test images come from an unseen physical sample (KTH-TIPS2) or add noise or an illuminant absent from the training images (TC32 and TC34).
The jointly learned components can therefore fit properties of the training images that do not hold at test time.
A learned color transform, in particular, mixes the channels, so even its intra-channel comparisons become partly inter-channel ones, the comparisons that an illuminant change flips (Section~\ref{sec:exp:photometric}).

\paragraph{Invariance by design, adaptation by learning}
\label{sec:analysis:invariance}
Learning adapts the operator to the training images but cannot create an invariance to a change that they do not contain.
An invariance must come from the operator or its input (median sampling in MRELBP, luminance input in gray-level descriptors) or from training data that contain the change.
The color pairs of FLEX-LBP are deliberately not invariant: the inter-channel contrasts that make it strong on natural color textures and histology (Sections~\ref{sec:exp:baselines} and~\ref{sec:exp:kather}) are what an illuminant change perturbs.
The remedies of Section~\ref{sec:exp:photometric}, like the image-relative magnitude code, are therefore choices of invariance rather than corrections of the learning, to be made from the nuisance expected at deployment.

\paragraph{Readout gap}
\label{sec:analysis:mismatch}
On NewBarkTex, TC13, Parquet and USPtex, LDA exceeds 1-NN for all three descriptors of Table~\ref{tab:endtoend}, by up to 23.34\,pp, so standardized 1-NN leaves discriminative information unused.
Because the untrained descriptor shows the same gap, it reflects mainly the readout and not the training objective, and vanilla NCA does not close it either (Supplementary Section~S3).
Learning narrows it on TC13, Parquet and USPtex, but the proxy loss favors linearly separable class prototypes, whereas 1-NN rewards low intra-class variance.
On KTH-TIPS2 and the suites with a controlled train/test change, the difference is near zero or negative for Fixed and more negative for Geometry (Supplementary Section~S20).
This is consistent with a decision boundary fitted under the training conditions.

With LDA on both descriptors, the gain of Geometry shrinks from $+3.55$ to $+0.92$\,pp on average, and that of All joint, paired over its three seeds, from $+3.69$ to $+2.68$\,pp (Section~\ref{sec:exp:postprocessing}).
Fixed read out by LDA is more accurate than Geometry read out by 1-NN, on average (86.52\% against 83.10\%) and on each of NewBarkTex, TC13, Parquet and USPtex.
With LDA on both sides, the gain of All joint over Fixed is significant in 10 of its 18 seed-level tests, against only 6 of 30 for Geometry.
These results suggest that learning the geometry mainly improves the neighborhood structure that nearest-neighbor matching uses directly.
On natural color textures, a supervised linear readout appears to recover much of the same separability from the fixed descriptor.
Learning the color transform as well adds accuracy that the linear readout does not recover, mainly on Parquet, NewBarkTex and TC31, but costs accuracy under the photometric shifts.

\subsection{Limitations}
\label{sec:analysis:limitations}

\paragraph{Comparisons}
We do not rerun the adaptive SPALBP~\citep{hu2024spalbp} and SARLBP~\citep{upadhyay2025sarlbp}, for which we found no public implementation.
Nor do we rerun CNN-based texture classifiers (Deep TEN~\citep{zhang2017deepten}, bilinear CNNs~\citep{lin2015bilinear}) or backbone-based encoders (RADAM~\citep{scabini2023radam}).
We do not fine-tune ResNet-50, although several of our training sets are large enough for fine-tuning to be common practice.
We also do not encode its intermediate layers with Fisher vectors~\citep{cimpoi2016deep} or try larger, newer or transformer backbones.
The frozen features are thus a lower bound, and fine-tuned models can be expected to keep an advantage that this paper does not aim to overturn.
NLBP, designed as a layer for deep backbones, is measured standalone, as a learned LBP descriptor.

\paragraph{Seeds, splits and test-split choices}
The four learned configurations other than Geometry and All joint use one seed, so their 1--2\,pp differences should be read with caution.
Outex, KTH-TIPS2 and NewBarkTex define a single official split; three random splits of Parquet and USPtex confirm the main conclusions on Fixed and All joint (Supplementary Section~S5).
The recommended Geometry configuration was identified from the seed-42 ablation, so only the additional seeds and Kather-CRC test it independently.
The radius templates were set per dataset with test results visible.
Swapping them changes the absolute accuracy by up to 4.4\,pp but not the sign of the learning gain (Supplementary Section~S16).
TC32 and TC34 were separated from the main benchmarks after their diagnosis but appear in every table.

\paragraph{Photometric invariance}
FLEX-LBP has no built-in invariance to illuminant color or to noise larger than the texture contrast, and each remedy costs accuracy where its nuisance is absent.
The remedies were found by diagnosing the two test suites, so their accuracies are not held-out results; PCANet remains ahead on TC34.

\paragraph{Histology protocol}
Kather-CRC is a single-center benchmark, evaluated as in its original study with tile-level folds that share slides between training and test.
Our histology results therefore do not measure generalization to new patients, centers or stains, which the multicenter follow-up dataset of \citet{kather2019survival} would allow.

\paragraph{The magnitude code is image-relative}
Our image-relative magnitude code discards the absolute size of the differences, a strong class cue on TC31 and Parquet.
There, fixed LTP coding is more accurate than Geometry, significantly in some seeds (Section~\ref{sec:exp:baselines}).
Recoding the trained Geometry model with an absolute threshold, without retraining, raises Parquet and TC31 but lowers five of the eight benchmarks (Supplementary Section~S17).

\section{Conclusion}
\label{sec:conclusion}

We present FLEX-LBP, an LBP whose color transform, radii, angles, thresholds and quantization can be learned by backpropagation, while straight-through estimation keeps every forward pass a hard-binary LBP.

(Q1)~Which components are worth learning depends on the dataset, and the best configuration of each benchmark can only be identified on its test split.
One configuration is nevertheless robust: learning the geometry and the thresholds with the color transform frozen.
Over five seeds, it improves the 1-NN accuracy of the identical fixed descriptor by $+3.55$\,pp on average, significantly on five of the six main benchmarks and never significantly worse.
It also improves the 1-NN accuracy of the fixed descriptor on colorectal histology ($+1.10$\,pp), a benchmark added after this configuration was chosen.
Learning everything can gain more on small, specialized datasets (Parquet, $+15.50$\,pp over three seeds) but, with 1-NN, loses consistently on KTH-TIPS2, and the color transform learned alone is never best.

(Q2)~Postprocessing adds $+2.02$\,pp to the fixed descriptor on average over the six main benchmarks, and the learning gain persists after it.
Fixed and learned descriptors can therefore only be compared fairly under identical postprocessing.
The readout matters as much: with a shrinkage LDA instead of 1-NN, the gain of Geometry falls to $+0.92$\,pp on average and that of All joint to $+2.68$\,pp.
LDA lifts the fixed descriptor above the learned geometry read out by 1-NN on four benchmarks and lowers the accuracy of the learned geometry on KTH-TIPS2 and TC31.
This suggests that learning the geometry mainly improves the neighborhood structure on which nearest-neighbor matching relies.

(Q3)~The learned operator remains interpretable: its radii, angles, thresholds and color transforms are explicit, bounded or regularized parameters.
Their direction of adaptation, such as a coarse radius that grows on one training set and shrinks on others, recurs across seeds and folds.

Within the LBP family, FLEX-LBP is more accurate than five classical descriptors, PCANet and NLBP on five of the six main benchmarks.
With LDA on every descriptor, it is again ahead on five of the six, but MRELBP then leads on KTH-TIPS2.
On colorectal histology, the fully learned descriptor reaches 93.82\%, significantly above the best classical descriptor (CLBP, 92.80\%), and 94.50\% with LDA (MRELBP, 90.44\%).
A fixed LTP coding matches Geometry on average, and most of its advantage over our code comes from its absolute dead zone.
This dead zone keeps the absolute size of the differences, mostly color differences between channels, that our image-relative magnitude code discards.
Under additive noise and illuminant change (TC32 and TC34), single-pixel comparisons drown and comparisons between channels flip.
Input-side remedies restore 96--97\%, level with BSIF (PCANet stays ahead on TC34), but each costs accuracy where its nuisance is absent.

Frozen ResNet-50 features (23.5 million parameters) are less accurate than Geometry on average with 1-NN and similar with LDA.
With LDA, All joint has the highest mean (89.07\%), a lead that rests on Parquet.
They are more accurate on KTH-TIPS2, USPtex and, with LDA, NewBarkTex, and less accurate on TC31, Parquet and, with 1-NN, histology.
They are also more robust to the photometric shifts without input remedies but fall below the best remedied FLEX-LBP descriptor with them.

In practice, without a reliable validation set, we recommend the Geometry configuration.
The color transform should be learned jointly with it only where held-out data can confirm the gain.
With an LDA readout, this joint learning falls below the fixed descriptor under the photometric shifts.
The input normalization should follow the nuisance expected at deployment (Section~\ref{sec:analysis:invariance}).
A per-channel (gray-world) normalization suits a possible change of illuminant or stain, and a median sampling support suits noise that may exceed the texture contrast.
When color and fine detail are reliable, neither remedy is needed.

Four directions follow from the limitations (Section~\ref{sec:analysis:limitations}; Supplementary Section~S21): (i)~photometric robustness that is not chosen by hand; (ii)~a magnitude threshold that interpolates between image-relative and absolute forms with a learned weight; (iii)~an objective or readout that uses what 1-NN leaves unused; and (iv)~a histology evaluation with slide-level splits and external centers~\citep{kather2019survival}, and a comparison with fine-tuned and larger deep texture models.

\section*{CRediT authorship contribution statement}
\textbf{Mohamed Alimoussa:} Conceptualization, Methodology, Software, Validation, Formal analysis, Investigation, Data curation, Writing -- original draft, Writing -- review \& editing, Visualization. \textbf{Amal El Kaid:} Writing -- review \& editing.

\section*{Declaration of competing interest}
The authors declare that they have no known competing financial interests or personal relationships that could have appeared to influence the work reported in this paper.

\section*{Funding}
This research did not receive any specific grant from funding agencies in the public, commercial, or not-for-profit sectors.

\section*{Data availability}
All benchmarks used in this study are publicly available (NewBarkTex, KTH-TIPS2-b, the Outex test suites, Parquet, USPtex and the colorectal cancer histology tiles of Kather et al., Zenodo record 53169, CC BY 4.0). The code of FLEX-LBP is available at \codeurl: a drop-in PyTorch layer with the training recipe, the postprocessing and the readouts.

\section*{Declaration of generative AI and AI-assisted technologies in the writing process}
During the preparation of this work the authors used Claude (Anthropic) to revise the text and check the manuscript against the code and data. After using this tool, the authors reviewed and edited the content as needed and take full responsibility for the content of the publication.

\bibliographystyle{elsarticle-harv}
\bibliography{refs_v22}

\end{document}